\documentclass[journal]{IEEEtran}

\usepackage{cite}

\usepackage{siunitx} 

\ifCLASSINFOpdf
   \usepackage[pdftex]{graphicx}
\else
\fi
\usepackage{amsmath}
\usepackage{amsfonts} 
\usepackage{array}

\ifCLASSOPTIONcompsoc
 \usepackage[caption=false,font=normalsize,labelfont=sf,textfont=sf]{subfig}
\else
 \usepackage[caption=false,font=footnotesize]{subfig}
\fi
\usepackage{mathtools}

\begin{document}
\title{High-Resolution Bathymetric Reconstruction From Sidescan Sonar With Deep Neural Networks}

%
%
%

\author{Yiping Xie,
        Nils Bore
        and~John~Folkesson
\thanks{This work was partially supported by the Wallenberg AI, Autonomous Systems and Software Program (WASP) funded by the Knut and Alice Wallenberg Foundation and partially supported by Stiftelsen för Strategisk Forskning
(SSF) through the Swedish Maritime Robotics Centre (SMaRC)
(IRC15-0046). (Corresponding author: Yiping Xie.)}
\thanks{The authors are with the Robotics, Perception and Learning Lab, Royal Institute of Technology, SE-100 44, Stockholm, Sweden (e-mail: yipingx@kth.se; nbore@kth.se; johnf@kth.se).}
}

\maketitle

\begin{abstract}
We propose a novel data-driven approach for high-resolution bathymetric reconstruction from sidescan. Sidescan sonar (SSS) intensities as a function of range do contain some information about the slope of the seabed. However, that information must be inferred.  Additionally, the navigation system provides the estimated trajectory, and normally the altitude along this trajectory is also available.  From these we obtain a very coarse seabed bathymetry as an input.  This is then combined with the indirect but high-resolution seabed slope information from the sidescan to estimate the full bathymetry. This sparse depth could be acquired by single-beam echo sounder, Doppler Velocity Log (DVL), other bottom tracking sensors or bottom tracking algorithm from sidescan itself.
In our work, a fully convolutional network is used to estimate the depth contour and its aleatoric uncertainty from the sidescan images and sparse depth in an end-to-end fashion. The estimated depth is then used together with the range to calculate the point's 3D location on the seafloor. A high-quality bathymetric map can be reconstructed after fusing the depth predictions and the corresponding confidence measures from the neural networks. We show the improvement of the bathymetric map gained by using sparse depths with sidescan over estimates with sidescan alone.  We also show the benefit of confidence weighting when fusing multiple bathymetric estimates into a single map.  
\end{abstract}



%
\IEEEpeerreviewmaketitle

\section{Introduction}
%
%
%


\IEEEPARstart{S}{idescan} and multibeam echo sounder (MBES) are the commonly used sonars for surveying the seabed. Sidescan sonar is used for obtaining detailed seabed images due to its high resolution and wide coverage, while MBES is used when constructing a bathymetric map due to its ability to directly measure the seafloor's 3D geometry.  The MBES are normally mounted on ships or large autonomous underwater vehicles (AUVs). Ones small enough for smaller AUVs will not have sufficient resolution in the across track direction.  They are also relatively expensive compared to single array sidescan sonars.  Since sidescan do not have any across track array, they can easily be mounted on small and more affordable AUVs.  The disadvantage is that there is also no across track angular resolution and thus the sidescan gives a 2D projection of the 3D seabed.    An estimate of the depth coordinates would resolve the 3D positions. 

\begin{figure*}[t]
\centering
\includegraphics[width=7.0in]{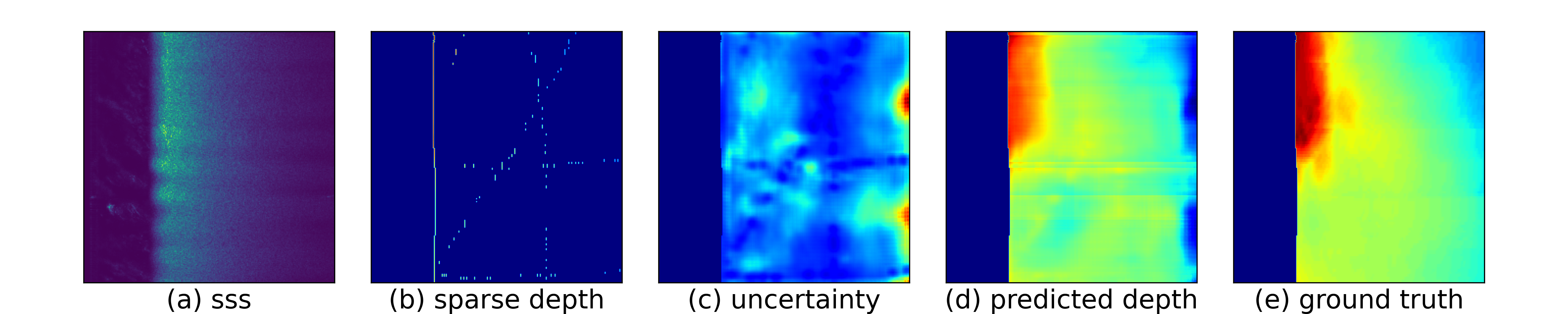}
\caption{An example of training image pairs. We divide the down-sampled sidescan waterfall image into smaller windows in our network training step and associate the depth to each pixel given bathymetry from the multibeam survey to form the ground truth. (a) Sidescan intensities input window with size $256\times256$. (b) Sparse depth input window. (c) Uncertainty estimation output windows. (d) Predicted depth output window. (e) Ground truth depth window.}
\label{fig:windows}
\end{figure*}
The sidescan intensities do contain some information about the seafloor's material and elevation changes: harder materials tend to have higher return intensities;  the nadir range in every ping gives a single altitude reading;  and, more importantly, the intensity changes indicate the change of the incidence angle \cite{folkesson20}. Even though the unknown bottom material and other disturbances make it difficult to estimate the depth from sidescan returns analytically, data-driven methods \cite{xie2019inferring} have shown promising results in estimating depth contours from sidescan intensities. Here we will add sparse height constraints from the  altimeter readings along the trajectory to the neural networks estimated depth approach.  This will significantly improve the accuracy of the method. 

In the last decades, deep learning has made a significant impact on the computer vision field. Among the various computer vision tasks, 3D reconstruction from monocular camera images can be seen as an analogous task to bathymetry from sidescan. Early on, shape from shading techniques based on physical principles were used in 3D scene reconstruction, but recently deep neural networks (DNNs) have become the state-of-the-art methods. Usually, DNNs estimate the depth from monocular images, and a pinhole model is used to reconstruct the 3D point clouds \cite{laina2016deeper}. 

Similar to how neural networks have been used for estimating depth from monocular camera images with sparse depth provided by low-resolution depth sensors like LiDARs \cite{ma2018sparse}, we train convolutional neural networks (CNNs) with sparse depth provided from the altimeter to predict a dense depth image from sidescan images, as in Fig. \ref{fig:windows}. 

 In theory, the sidescan intensities contain information on the surface gradients. If we integrate the surface gradients, it will inevitably drift further as we get far from the starting point at the nadir. Our prior work \cite{xie2019inferring} shows that the estimated errors are most significant as one moves further from the sensor. As a matter of fact, there are problems of treating sidescan images more or less as camera images in a CNN. A convolutional filter assumes that the interpretation is invariant to pixel position. For sidescan images, that is not exactly the case. There is a changing interpretation as one moves further away. The geometry shifts and the per column change rate of the incidence angle are less in the far than the close region. 
 
\begin{figure}[t]
\centering
\includegraphics[width=3.0in]{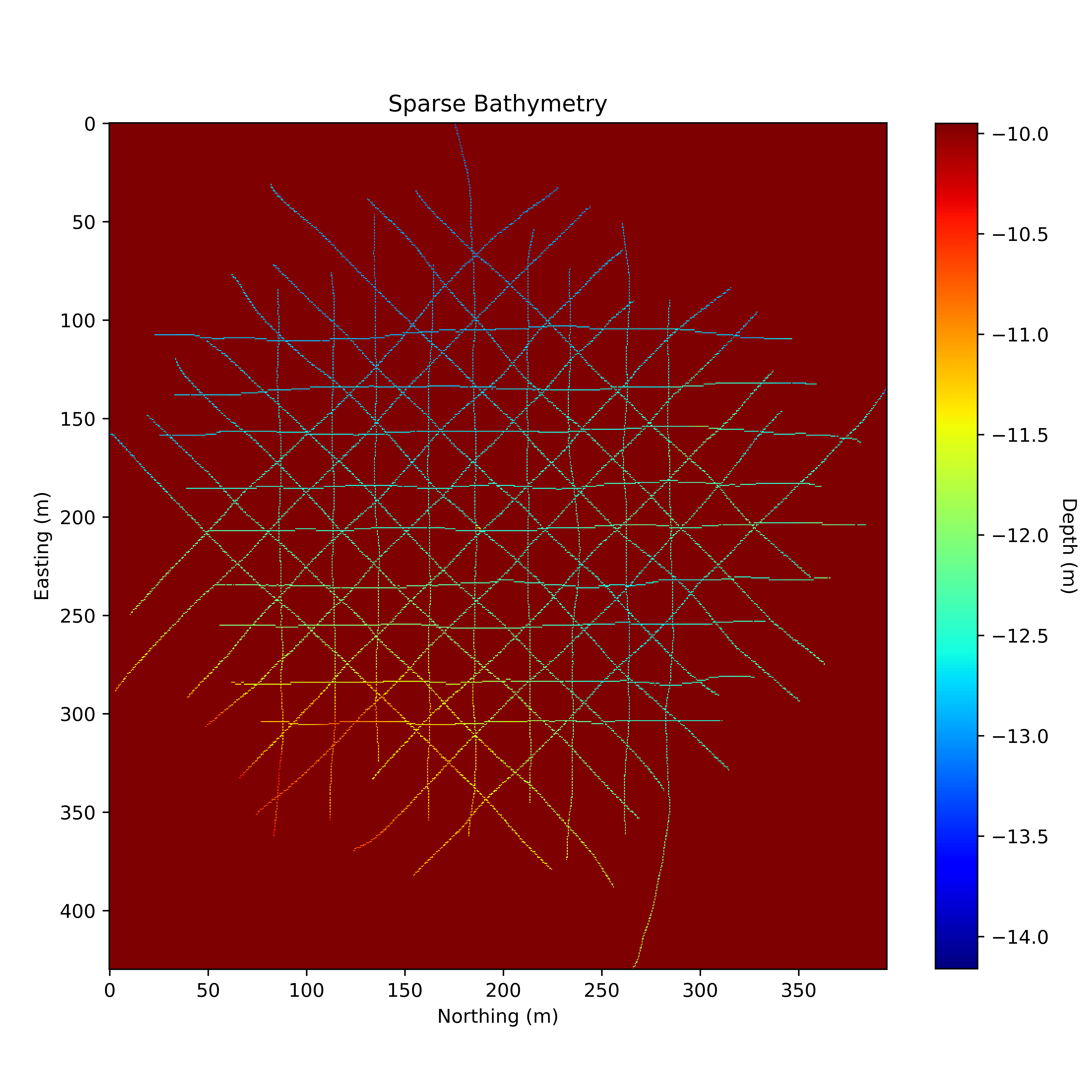}
\caption{An example of sparse seabed bathymetry from Dataset 2.  Brown represents no data and the colored tracks are the lines of sparse data.}
\label{fig:sparse_bathymetry}
\end{figure}

 To address the increasing errors further from the sidescan sensor, we utilize the sparse depth (see Fig. \ref{fig:sparse_bathymetry}) and an estimate of uncertainty to reconstruct a better bathymetry. We use the sparse depth provided by navigating and altitude measurement, as a constraint to the neural network, to reduce the drift errors. We use the uncertainty estimation to form a probabilistic model to fuse different estimations from different lines of the survey. 
 
Our contributions are that:
\begin{enumerate}[\IEEEsetlabelwidth{3)}]
\item We show that a novel framework to reconstruct bathymetry with high resolution and high quality with sidescan and sparse depth improves the accuracy over only using either of those two inputs;
\item  We show that an aleatoric uncertainty estimation as a confidence measure for the depth estimation can improve the bathymetric map formed by combining estimates from overlapping survey lines.
\end{enumerate}

\subsection{Related Work}

Woock and Frey \cite{woock2010deep} summarize the challenges of extracting depth information from sidescan sonar (SSS), which requires knowledge of sediment characteristics, surface and volume scattering properties, sound absorption and dispersion, water currents, variations in sound speed, and the sonar transducer beam pattern. Assumptions must be made to simplify the methods, such as isovelocity Sound Velocity Profile (SVP). 

Many attempts to reconstruct a shape from sidescan are based on "Lambert's cosine law". Li and Pai's work \cite{li1991improvement} is inspired by shape-from-shading methods with camera images \cite{zhang1999shape}, determining a Lambertian sonar model to obtain the approximation of the surface normals. However, the diffuse reflections assumptions work much better for light than for sound. Coiras \textit{et al.} \cite{coiras2007multiresolution} model the intensity as a function of bathymetry, the reflectance, and the incident energy. They model the bathymetry and reflectance by splines and the incident energy by polynomials to reduce the dimensionality and apply standard gradient descent to the square error in modelled intensity versus measured. In \cite{coiras2007multiresolution}, quality and quantity validation are done on a pipe of known diameter.

Jones and Traykovski \cite{jones2018method} collect data with a rotary sidescan sonar, which is mounted on underwater frames and rotated $360^\circ$ to get a circular image. They exploit the shadows in sidescan images to estimate the elevation of bedform in shallow water and validate their methods on wave-orbital ripples and megaripples comparing with the multibeam data.
Usually sidescan data are collected with overlapping swaths and this overlap can be used to infer depth from the images. Burguera and Oliver \cite{burguera2016high} exploit a physics-based SSS model to correct the raw data including beam corrections and motion estimation, leading to a probabilistic framework to build a high-resolution bathymetric map from sidescan data. 
Zhao \textit{et al.} \cite{zhao2018reconstructing} also integrate the sparse bathymetry into the reconstruction, utilizing a bottom tracking algorithm as an altimeter to obtain an initial seabed topography, which is then used as a constraint for the reconstruction model based on Lambertian law. The evaluation is done compared to the bathymetry constructed by a single-beam bathymetric system.

Deep learning approaches have been used in sidescan images for other tasks in recent years, such as object classification \cite{dzieciuch2016non}, \cite{huo2020underwater}, object detection \cite{einsidler2018deep}, and semantic segmentation \cite{rahnemoonfar2019semantic}, \cite{wu2019ecnet}. 
Dzieciuch \textit{et al.} \cite{dzieciuch2016non} show that a simple CNN can be used for mine detection in sidescan sonar imagery and achieve comparable accuracy as human operators. Huo \textit{et al.} \cite{huo2020underwater} show that with deep transfer learning, a CNN could achieve high accuracy on the multi-class classification task on sidescan images. They also propose a semi-synthetic data generation method to handle the imbalanced training data, which is a most common case in the real sidescan datasets. 
Einsidler \textit{et al.} \cite{einsidler2018deep} show that deep transfer learning could also be used for underwater object detection. They adapt the state-of-the-art object detection algorithm, YOLO (You Only Look Once) \cite{redmon2016you}, to sidescan images, and achieve reasonable accuracy in anomaly detection after some fine-tuning on the real sidescan dataset.
Rahnemoonfar and Dobbs \cite{rahnemoonfar2019semantic} propose a novel CNN architecture and illustrate its performance on pothhole semantic segmentation of sidescan images. Wu \textit{et al.} \cite{wu2019ecnet} propose ECNet to perform semantic segmentation on sidescan with much fewer parameters and much faster speed, making it possible to be applied to real-time tasks on embedded platforms.

Our previous work \cite{xie2019inferring} show promising results for the task of depth estimation from sidescan images with deep learning techniques. 
Inspired by deep learning methods to estimate depth from single camera images \cite{liu2015deep}, in \cite{xie2019inferring}, we propose a method to extract 3D information from 2D sonar images with DNNs. In this work, based on our prior one, we further exploit the sparse depth as a constraint for the DNNs and propose a framework of building a complete bathymetric map from sidescan. 
The use of the sparse depth is inspired by \cite{ma2018sparse}, namely, a deep regression network taking the sidescan and sparse depth data as input.  In \cite{ma2018sparse} they demonstrate their proposed framework outperforms the other depth fusion techniques.
The uncertainty estimation in this work is closely related to \cite{lakshminarayanan2016simple} and \cite{walz2020uncertainty}. In \cite{lakshminarayanan2016simple}, they propose a simple framework to quantify predictive uncertainty in neural networks, which is easy to adapt to most of the deep learning approaches. They show that maximizing likelihood is a proper scoring rule, which measures the quality of predictive uncertainty \cite{gneiting2007strictly}. In \cite{walz2020uncertainty}, they demonstrate the framework proposed by \cite{lakshminarayanan2016simple} can also achieve reasonable results on pixel-level applications such as depth estimation.  And simply by filtering out the few extreme outliers with high uncertainty, one can improve the overall performance of the 3D reconstruction.

The major difference between our work and the others to reconstruct bathymetry from sidescan is that we use a data-driven approach whereas the prior work are model-based. Our motivation is that some effects are not plausible to model yet the sidescan images do contain some information about them. For example, an expert can tell if a sidescan image appears to be hard or soft bottom, rocks appear often as part of a larger geological formation and so on. It is not practical, however, to have experts estimate all the sediment characteristics in every sidescan images, and model the surface scattering properties accordingly. Thus, we exploit data-driven methods to leverage deep learning's advantages of learning from patterns in the data distributions to compensate for those unmodelled effects.

\subsection{Summary of the Proposed Method}
Reconstructing the bathymetry from sidescan sonar is difficult. Many properties that are hard to model have large impacts on estimating depth contours from sidescan. With a data-driven approach, some of these can be partially compensated, but naturally there will be errors.  Besides the unmodeled properties of the seabed and water column, the main  source of errors is the navigation error between lines of the survey that provide the sparse depth information. 

In this article, we develop a method that reconstructs the bathymetry relying on sidescan sonar, vehicle position, and the altimeter. Such a data-driven method could in principle work with data produced by most
standard sidescan surveys. We utilize the sparse depth to reduce the errors and propose a framework to estimate the depth and uncertainty at the same time, and a probabilistic model to reconstruct the bathymetry.

\section{Method}
\subsection{Sidescan Sonar Formation}
\begin{figure}[t]
\centering
\includegraphics[width=3.5in]{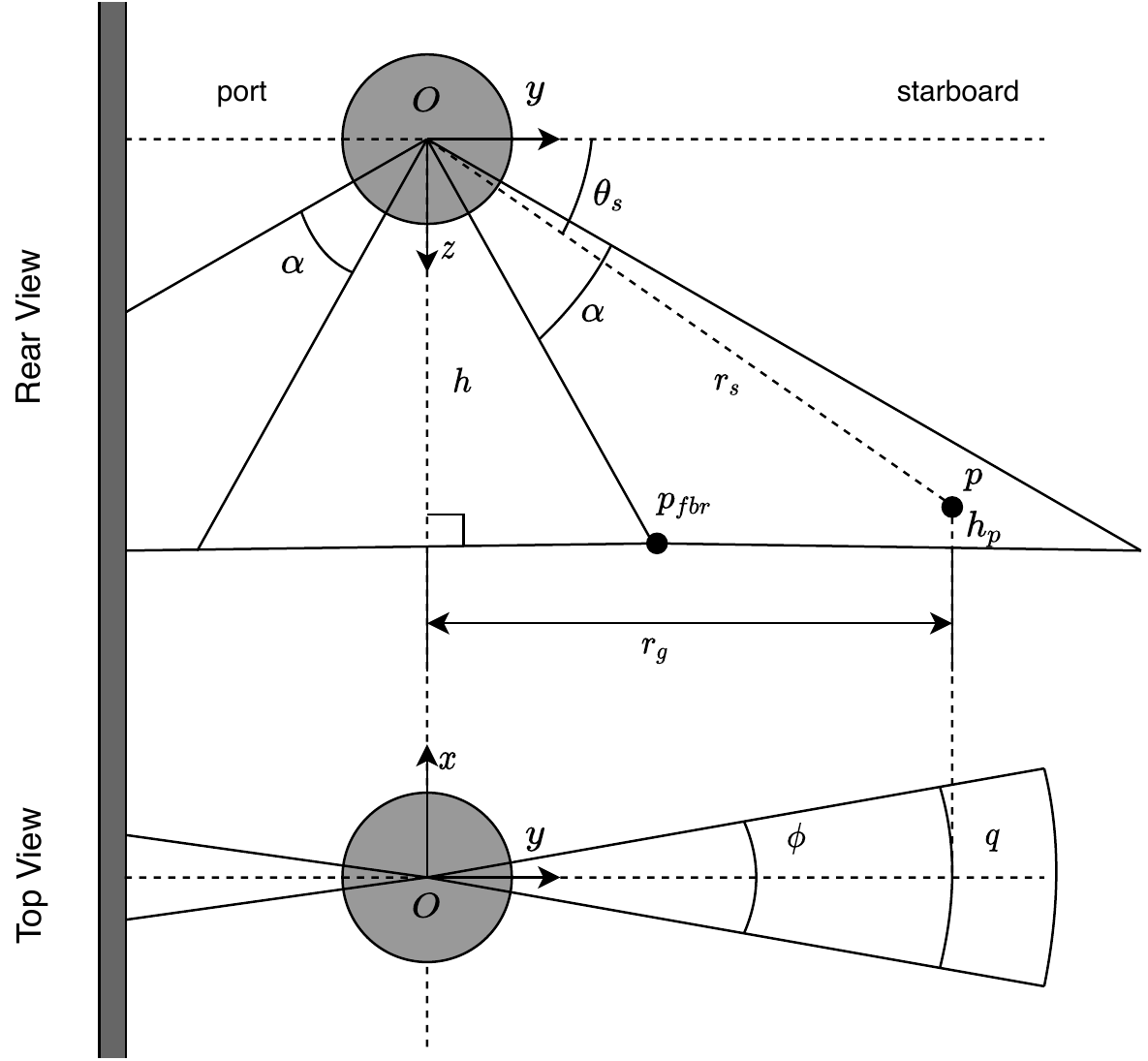}
\caption{Sidescan sonar formation.}
\label{fig:sss-formation}
\end{figure}

Fig. \ref{fig:sss-formation} illustrates the top view and the rear view of a sidescan sonar with its sensor origin $O$ at altitude $h$. Let $p$ be a point in the ensonified region on the bathymetric surface $\mathcal{M}\subset \mathbb{R}^3$  with point altitude $h_p$, whose polar coordinates can be expressed in its slant range $r_s$ and its grazing angle $\theta_s$. The grazing angle $\theta_s$ can be calculated as follows if we ignore ray bending effects and $h_p$ is known:
\begin{equation}\label{eq:grazing_angle}
    \theta_s = \arcsin{\Big(\frac{h-h_p}{r_s}\Big)}.
\end{equation}
The ground range $r_g$ is the projection of vector $\Vec{op}$ over the $Y$ axis. The vertical beam width $\alpha$, sometimes referred to as sensor opening in the $YZ$ plane, is usually $40\text{-}60^\circ$, and the horizontal beam width $\phi$, sometimes referred to as sensor opening in the $XY$ plane, is usually around $0.1^\circ$ \cite{blondel2010handbook}. Due to the horizontal beam width $\phi$, the exact point position of $p$ in the $XY$ plane is ambiguous over the arc $q$; however, the assumption is usually that this fact can be neglected since $\phi$ is very small. 

\subsection{Sparse Depth Association}
The idea is to use the set of points directly below the AUV along with its altitude and pose reading to generate a set of sparse depths, $x, y, z$.  Then for a waterfall image from any line in our survey we can compute the range from the sonar at each ping to points from the sparse depth set that fall within its range and beam angle.  We then create a second sparse depth image where the pixels correspond to the sidescan waterfall image but the values are now depths relative to the depth of the sonar [see Fig. 5(b)].

\subsection{Uncertainty Estimation}
Predicting depth from sidescan images can be seen as a pixel-level regression problem that can be addressed using neural networks.  We pre-process the data so that the network is trying to estimate the point altitude $h_p$. 
We set the final layers of the neural network to output two values: mean $\mu(h_p)$ and variance $\sigma^2(h_p)$.  We do a variational fit of the point altitude to a Laplacian distribution with the predicted mean $\mu(h_p)$ and variance\footnote{Similar to \cite{lakshminarayanan2016simple}, we enforce the positivity constraint on $\sigma^2(h_p)$ by passing it through the softplus function, $\log(1+\exp(\cdot))$ and add a minimum constant e.g., $10^{-6}$ for numerical stability.} $\sigma^2(h_p)$. The loss using negative log-likelihood (NLNL) will be:
\begin{align}
    -\log p_\theta (h_{p,gt,n}|d_n) =  \frac{\lVert h_{p,gt}-\mu_\theta(h_p)\rVert}{\sigma^2_\theta(h_p)} + \log \sigma^2_\theta(h_p) + \log2.
\end{align}

As a comparison, the mean absolute error, MAE, loss can be seen as a special case of minimizing the above loss with a constant variance $\sigma^2_\theta=1$\footnote{The value of 1 is arbitrary but other choices would only scale the loss and change the constant part and thus have no effect on the optimization.}. We model the likelihood to follow Laplacian distribution instead of Gaussian because we find $L1$ loss is more suitable than $L2$ loss for depth regression, as observed in  \cite{walz2020uncertainty}. 
Therefore, the NLNL averaged over each pixel in sidescan images is considered as an aleatoric loss function for training the neural network \cite{walz2020uncertainty}.

During the test phase, we can use $c_p=\frac{1}{|\sigma^2_\theta(h_p)|}$ as a confidence measure of the depth estimates, where $\sigma^2_\theta(h_p)$ is the uncertainty output of the network. By fusing all confidence estimates together we are able to create a confidence map $\mathcal{U}\subset \mathbb{R}^3$ for the corresponding reconstructed bathymetry $\hat{\mathcal{M}}\subset  \mathbb{R}^3$.
\subsection{Bathymetry Reconstruction Model}\label{par:bathymetry_model}
For a sidescan waterfall image, let $I^{k,i}$ denote the returned intensity corresponding to ping number $k$ and echo travel time interval $i$, $r_s^{k,i}$ denote the corresponding slant range and $r_g^{k,i}$ denote the ground range. The slant range can be deduced from the sound speed $c^k$ and the two-way travel time $t^{k,i}$ between the sidescan sonar and the point at seafloor $\mathbf{p}^{k,i}$ as follows:
\begin{align}\label{eq:slant_range}
r_s^{k,i} &= \frac{c^k \cdot t^{k,i}}{2}.
\end{align}
Note here we assume an isovelocity SVP, which will introduce additional errors that could be eliminated if the SVP were known. 
The rotation  $\mathcal{R}^k\in SO(3)$ and position $\mathbf{s}^k$ of the sonar is given by the navigation, and the point altitude $h_p^{k,i}$ can be estimated from our neural network. Thus the point position $\mathbf{p}^{k,i}\in \mathbb{R}^3$ can be calculated by simply solving the following equation:

\begin{IEEEeqnarray}{rCl}\label{eq:op_norm2}
  r_s^{k,i} &=&\lVert \mathbf{s}^k - \mathbf{p}^{k,i} \rVert_2 \IEEEnonumber\\
          &=&\sqrt{(r_g^{k,i})^2+(h^k-h_p^{k,i})^2}.
\end{IEEEeqnarray}
 We can now fuse all estimates $\mathbf{p}^{k,i}=(p_x^{k,i},p_y^{k,i},p_z^{k,i})$ from the neural network from every survey line. We add them in a probabilistic fusion model to form a bathymetric mesh $\hat{\mathcal{M}}$ using the confidence estimates $c_{p}^{k,i}$. So a fused depth for point $p_z$ on the reconstructed bathymetry grid $\hat{\mathcal{M}}$
is:
\begin{equation}\label{equ:bathymetry_model}
    \hat{p}_z = \frac{\sum\limits_{p\in\mathcal{P}} p_{z}^{k,i}c_{p}^{k,i}}{\sum\limits_{p\in\mathcal{P}} c_{p}^{k,i}}
\end{equation}
where $\mathcal{P} \subset \mathbb{R}^3$ is the set of points that fall within the grid cell.  The fused confidence map $\mathcal{U}$ can be obtained by averaging $c_p^{k,i}$ over $\mathcal{P}$:
\begin{align}
    \hat{c} &= \frac{\sum\limits_{p\in\mathcal{P}} c_{p}^{k,i}}{|\mathcal{P}|}.
\end{align}

\begin{table}[t]
\renewcommand{\arraystretch}{1.3}
\caption{Datasets Details}
\label{tab:datasets}
\centering
\begin{tabular}{c||c|c}
\hline
 & Dataset 1 & Dataset 2\\
  & \multicolumn{1}{c|}{Training Validation Testing} & Testing \\
\hline
Survey lines & \multicolumn{1}{c|}{45 \;\;\;\;\;\;\;\;\; 6  \;\;\;\;\;\;\;\;\; 6} & 36\\
\hline
Image pairs & \multicolumn{1}{c|}{4352 \;\;\;\;\; 640  \;\;\;\;\; 592} & 1994\\
\hline
Max Depth & 25.07 m & 13.66 m\\
\hline
Min Depth & 9.03 m & 10.45 m\\
\hline
Sonar type & Edgetech 4200MP & Edgetech 4200MP\\
\hline
Sonar range & $\sim$50 m & $\sim$50 m\\
\hline
Sonar frequency & 850 kHz & 850 kHz\\
\hline
\end{tabular}
\end{table}

\subsection{Sidescan Draping and Dataset Generation}\label{par:sidescan_draping}
\subsubsection{Sidescan Geographic Referencing}
To generate ground truth for the training and validation datasets, we need to associate sidescan intensities $I^{k,i}$ to its georeferenced coordinates $\mathbf{p}^{k,i}$ on a bathymetric mesh $\mathcal{M}\subset \mathbb{R}^3$, which is also referred to as \textit{sidescan draping} \cite{Bore2020}. To do so, the MBES is used to form such mesh, and the SVP is needed to determine the sound speed of the water layer. Also, the sensor position $\mathbf{s}^k \in \mathbb{R}^3$ and the rotation matrix $\mathcal{R}^k\in SO(3)$ of the sidescan must be known. Using this we find the intersection of each sidescan sonar  arc with the mesh.

\subsubsection{Dataset Generation}
The two datasets (see Table \ref{tab:datasets}) we used in this paper were both collected with MMT Ping, a survey vessel equipped with a hull mounted sidescan Edgetech 4200MP and RTK GPS to ensure high accuracy positioning. For both datasets, we have the high-resolution multibeam bathymetry collected with Reson 7125, treated as the ground truth. For every survey line, we divide each side of the waterfall image into smaller windows with height $H=256$ and width $W=256$ downsampled from $\sim 6000$ bins. The selection of $H$ and $W$ is chosen to fit the convolutional neural networks and at the same time ensure the sidescan's across-track resolution higher than the bathymetry resolution.

\subsection{CNN Model}
The model takes the sidescan intensities window [see Fig. \ref{fig:windows}(a)] and sparse depth window [see Fig. \ref{fig:windows}(b)] directly concatenated together as the input and output the estimated depth and uncertainty. The loss function is the negative log-likelihood averaged over each valid pixel in the window:
\begin{align}
    \mathcal{L}=\frac{1}{|D_{k,i}|}\sum_{d \in D_{k,i}}  \Big(\frac{\lVert d_{gt}-\mu_\theta(d)\rVert}{\sigma^2_\theta(d)} +\log \sigma^2_\theta(d) \Big)
\end{align}
where $\sigma^2_\theta(d)$ is ensured to be positive and $D_{k,i}$ is the set of all valid depth points by masking out the nadir area and missing data.

\begin{figure}
\centering
\includegraphics[width=3.0in]{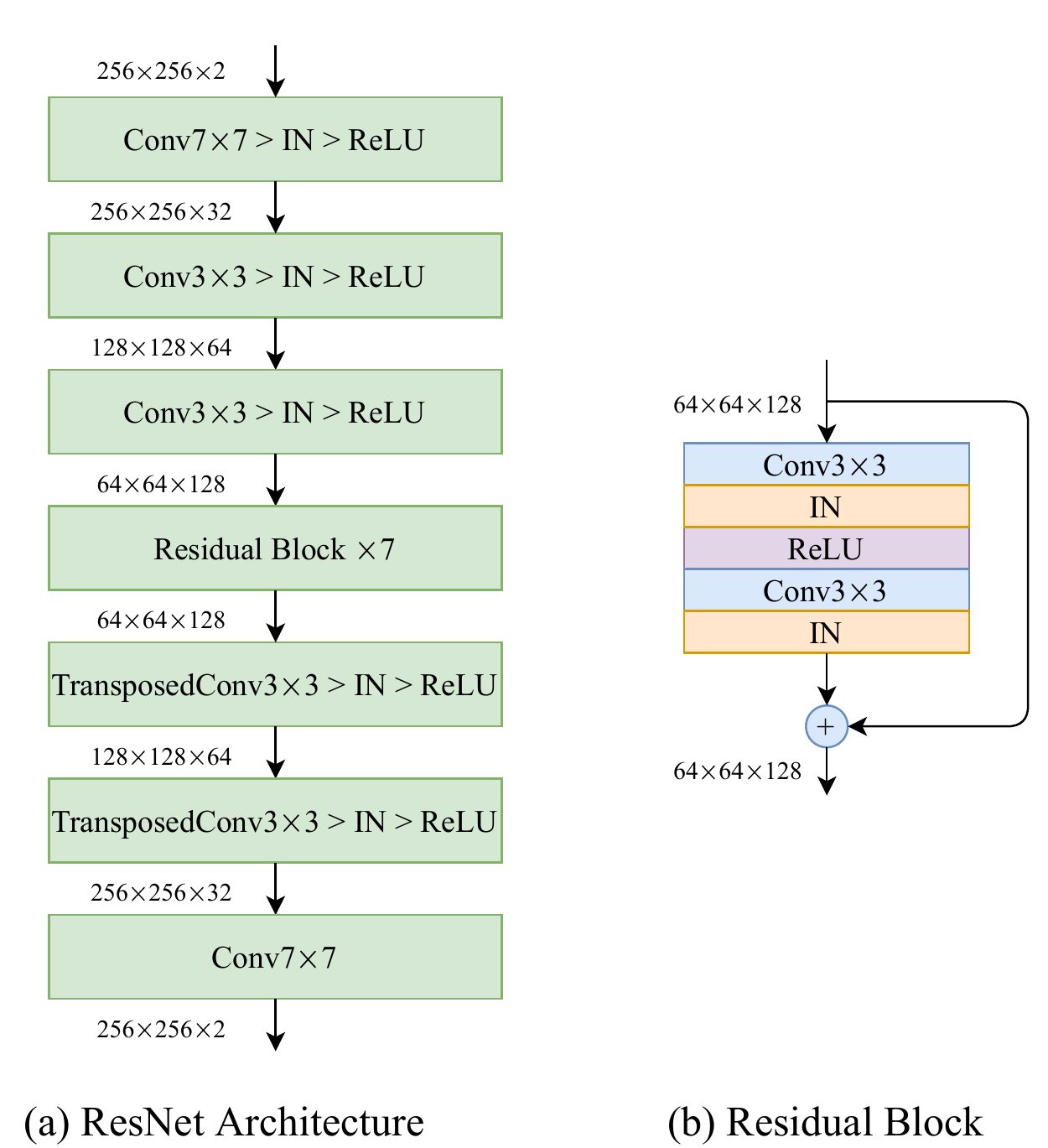}
\caption{CNN architecture. (a) Network architecture. (b) Resblock.}
\label{fig:resnet}
\end{figure}

The neural network architecture is a Fully Convolutional Network (FCN) based on our prior work \cite{xie2019inferring} with some minor modifications to adapt the sparse depth and uncertainty estimation, shown in Fig. \ref{fig:resnet}(a). For the normalization we choose Instance Normalization (IN) and for the activation functions we use Rectified Linear Unit (ReLU). The down-sampling layers consist of three convolutional modules in the form of convolution-IN-ReLU. These are followed by the seven residual blocks, shown in Fig. \ref{fig:resnet}(b).
The residual blocks consist of several layers with no change in the image dimension, the output of which is summed with the input and fed to the next residual block. By feeding the input directly to the output one gets a direct link across all the blocks that facilitate propagation of the gradient. The residual blocks are followed by two upsampling layers with two transposed convolution layers and the convolution operation in the end. 

\section{Experiments}

The method is evaluated on two datasets from different areas. Dataset 1 is divided into training, validation and test sets, while Dataset 2 is only used for testing. The details are shown in Table \ref{tab:datasets}.

\begin{figure*}[!t]
\centering
\subfloat[SSS Waterfall Image]{\includegraphics[width=2.1in]{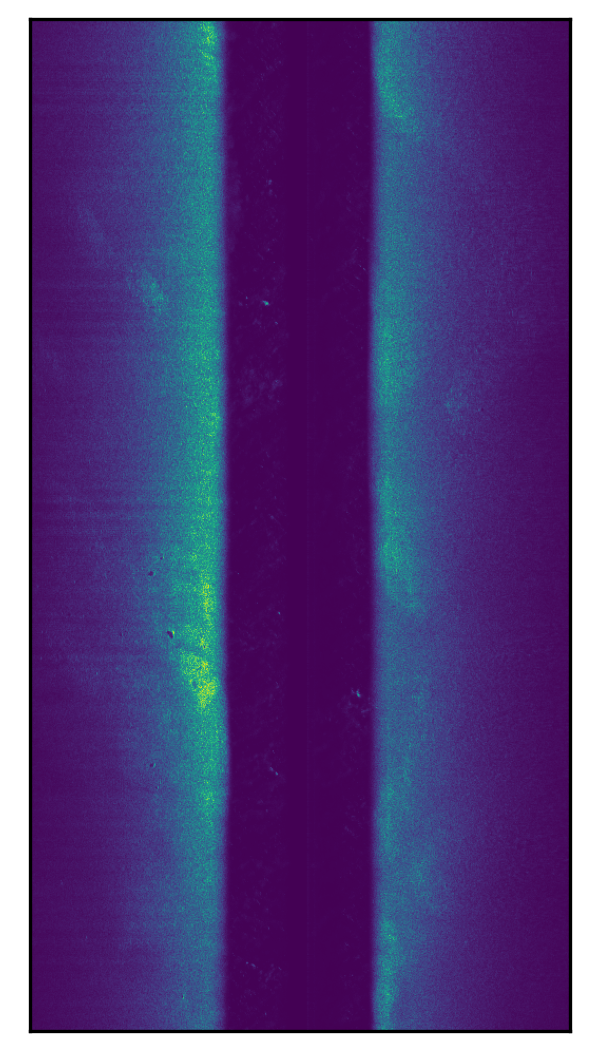}
\label{fig:sss_sparse_depth_a}}
\hfil
\subfloat[Sparse Depth Waterfall Image]{\includegraphics[width=2.1in]{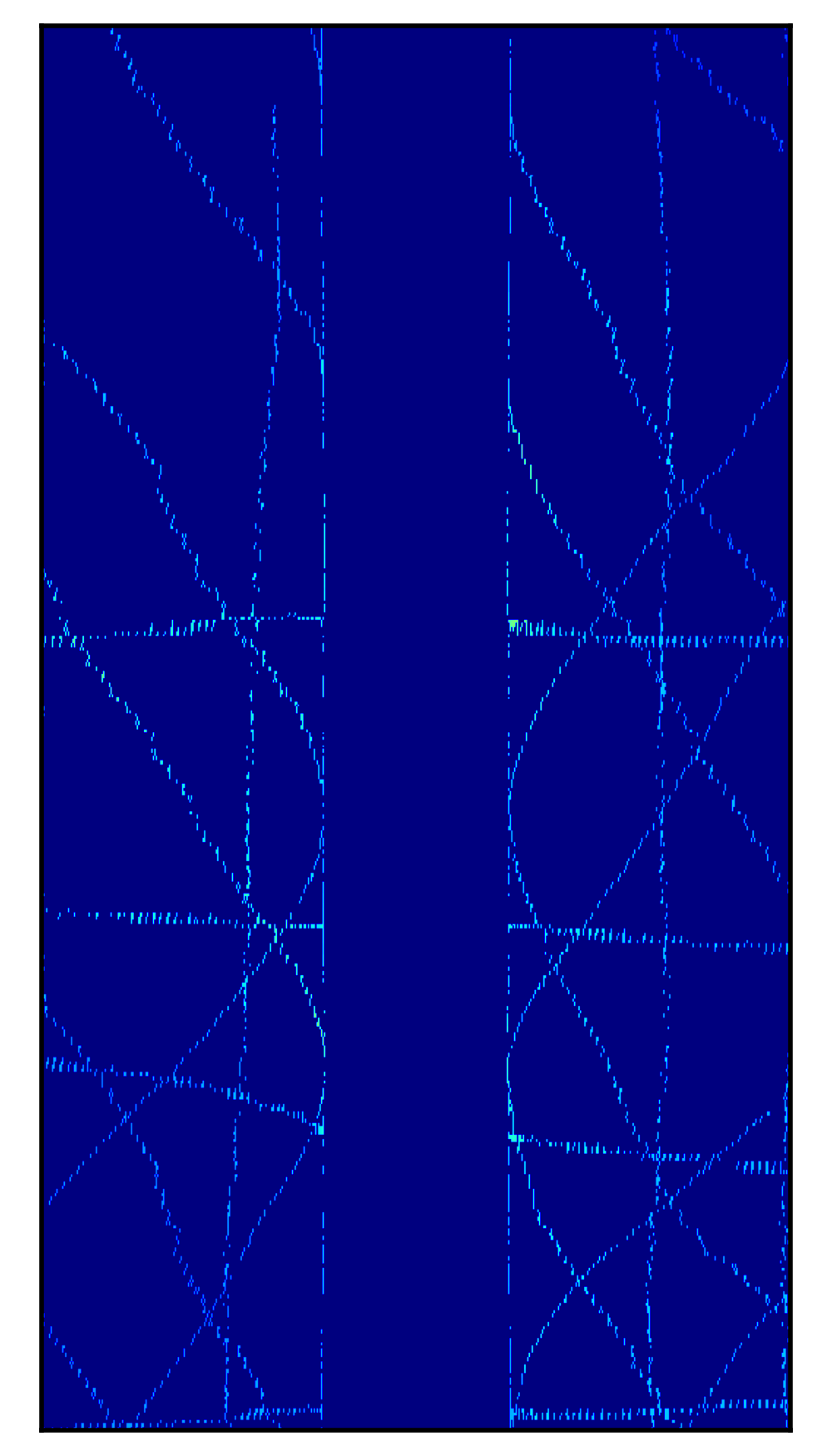}
\label{fig:sss_sparse_depth_b}}
\hfil
\subfloat[Uncertainty Waterfall Image]{\includegraphics[width=2.1in]{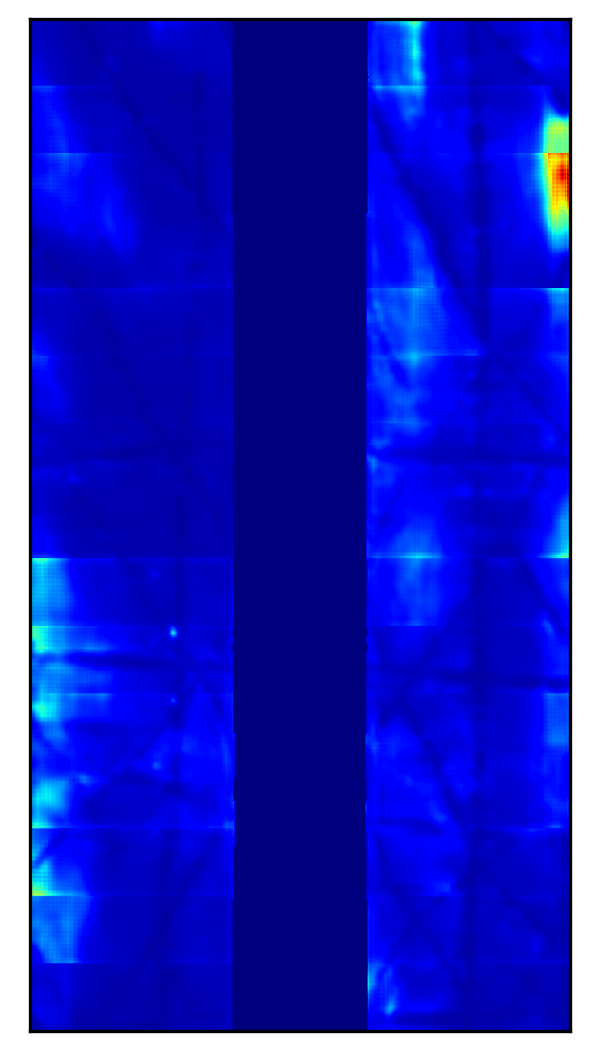}
\label{fig:sss_sparse_depth_c}}
\caption{An example of part of one sidescan waterfall image, the associated sparse depth and the estimated uncertainty from a survey line in Dataset 2. Rainbow colormap is used to show the uncertainty. We can observe that the uncertainty is low (dark blue) at pixels where sparse depth is available.  Also note that the uncertainty does not increase with distance from the nadir as would be the case without sparse depths.}
\label{fig:sss_sparse_depth}
\end{figure*}
Using \textit{sidescan draping} described in Section \ref{par:sidescan_draping}, we can associate the waterfall images to corresponding depth images and the sparse depth available from the altimeter reading along the trajectory, seen Fig. \ref{fig:sss_sparse_depth}. 
For each side of the waterfall image, we divide it into smaller windows with height $H=256$ and width $W=256$ downsampled from $\sim 6000$ bins. The square images with size $256\times256$ make it easier to adapt the architecture of the mainstream neural networks for computer vision. To generate more training data, we augment the data by allowing the windows to overlap by 75\% and flipping the windows in the along-track direction to simulate the sonar is moving to exactly the opposite direction.  

The network is trained on the training set with 4352 windows from Dataset 1, with different hyper-parameters. The validation set is used to select the three best models, whose results will be used for ensemble in the testing phase later to make better estimation of the predictive uncertainty from the neural network. The six lines from validation set, Dataset 1 and the six lines from test set, Dataset 1 are evenly distributed across the whole area but orthogonal to each other. The purpose of this is to test the generalization of the network when the sidescan images are from $90^\circ$ angles. The whole Dataset 2 from totally another place is also used as the test set, to test the generalization of the network when coming to different environments. 

To evaluate the methods, we compare the bathymetric map generated from the network and the one from the MBES pings. The bathymetric map is generated by solving the reverse problem of \textit{sidescan draping} with the methodology described in Section \ref{par:sidescan_draping}. Due to sidescan's wide coverage and high resolution, for most of the points on the seafloor, there are usually many estimates. During the bathymetry fusion, we first discard the extreme outliers with uncertainty larger than a certain threshold, and then use the confidence measurement $c_p^{k,i}$ as weights for the corresponding depth estimates, as \eqref{equ:bathymetry_model} in Section \ref{par:bathymetry_model}.

To compare the result, we use the same resolution (0.5 m) as the bathymetric map from the MBES data, where in theory the resolution is not integral to the method, meaning one could choose much higher resolution to build a super-resolution bathymetric map based on sidescan data. The potential challenge is the lack of super-resolution bathymetry to evaluate and the GPU power to train the network.

\section{Results}\label{par:results}
\subsection{Reconstruction Results} 
\begin{figure}[t]
\centering
\includegraphics[width=3in]{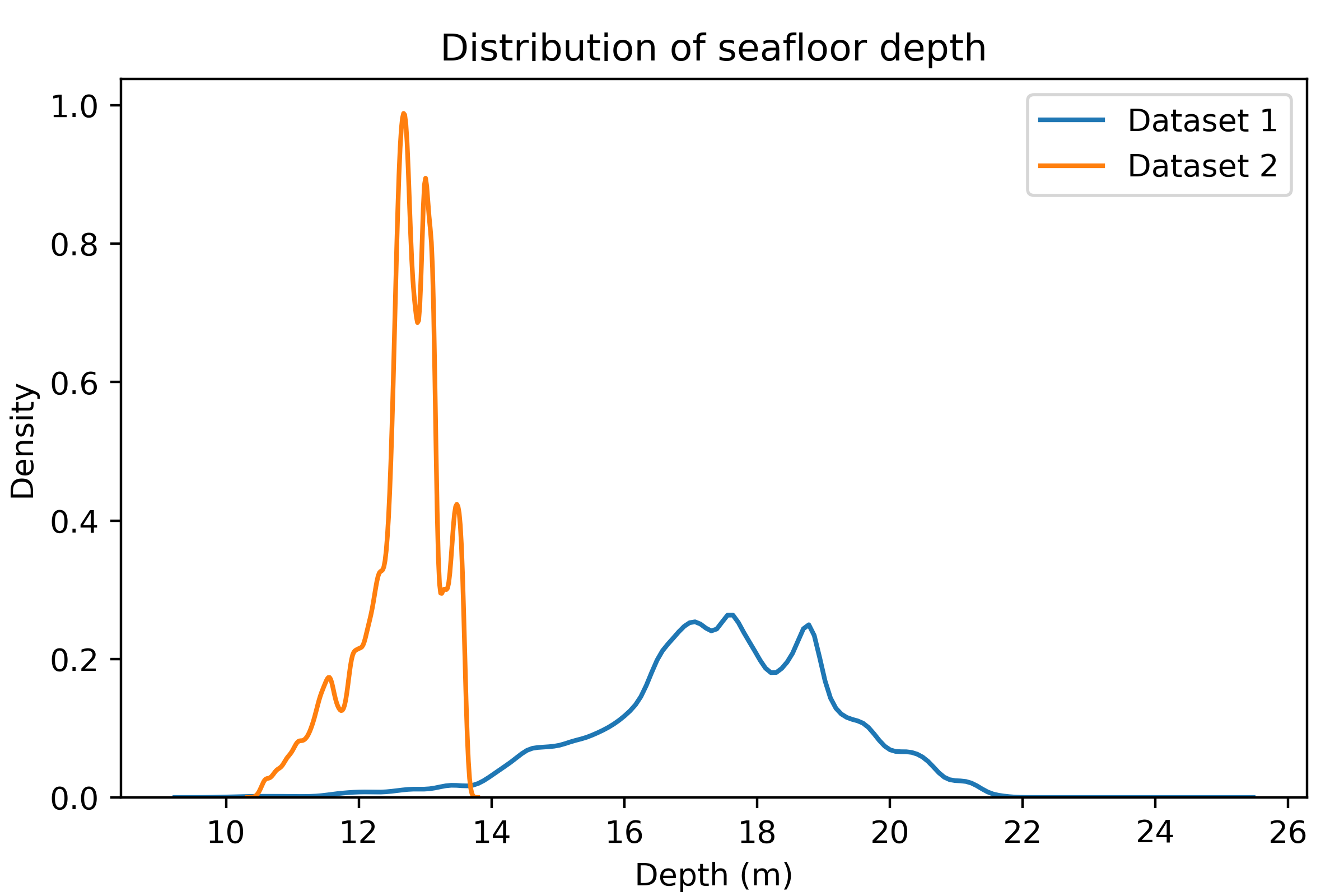}
\caption{Distributions of depth in two datasets.  The relative depth distribution would be the same offset by the nearly constant depth of the sonar. }
\label{fig:depth_density}
\end{figure}
Fig. \ref{fig:depth_density} shows the seafloor depth distribution of two datasets. we can notice that Dataset 1 covers a large range of 9-21 m, whereas Dataset 2  mainly concentrates on the range of 10-14 m.   
\subsubsection{Dataset 1}
\begin{figure}[t]
\centering
\includegraphics[width=3.5in]{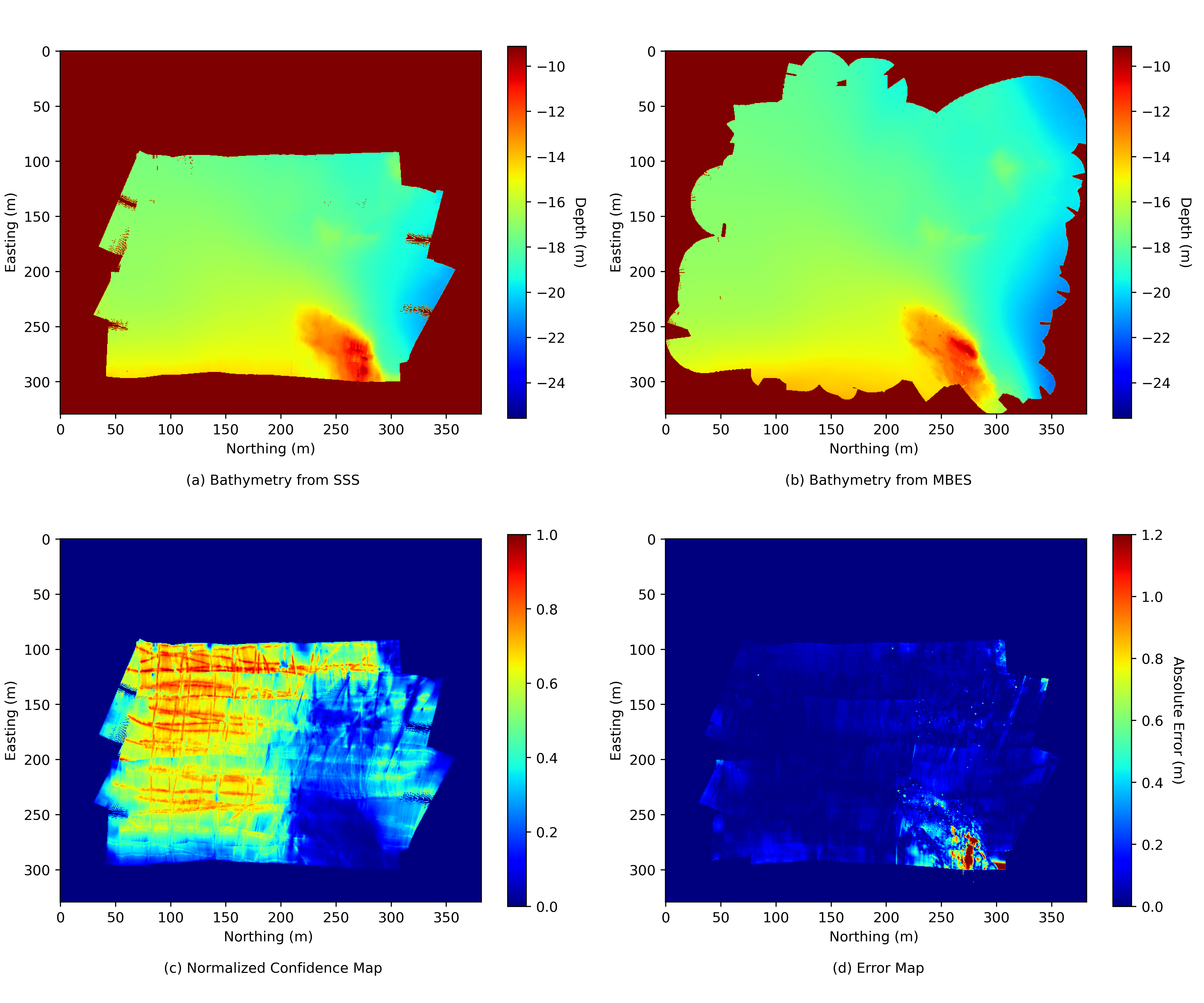}
\caption{Bathymetry on Dataset 1, produced by sidescan and multibeam respectively. (a) Bathymetry from six sidescan survey lines. (b) Ground truth bathymetry produced with multibeam data. (c) Normalized confidence map for the bathymetry produced from sidescan, color red indicating high confidence low uncertainty. (d) Absolute error map between (a) and (b).}
\label{fig:Goteborg_height_map}
\end{figure}

\begin{figure}[t]
\centering
\includegraphics[width=3.5in]{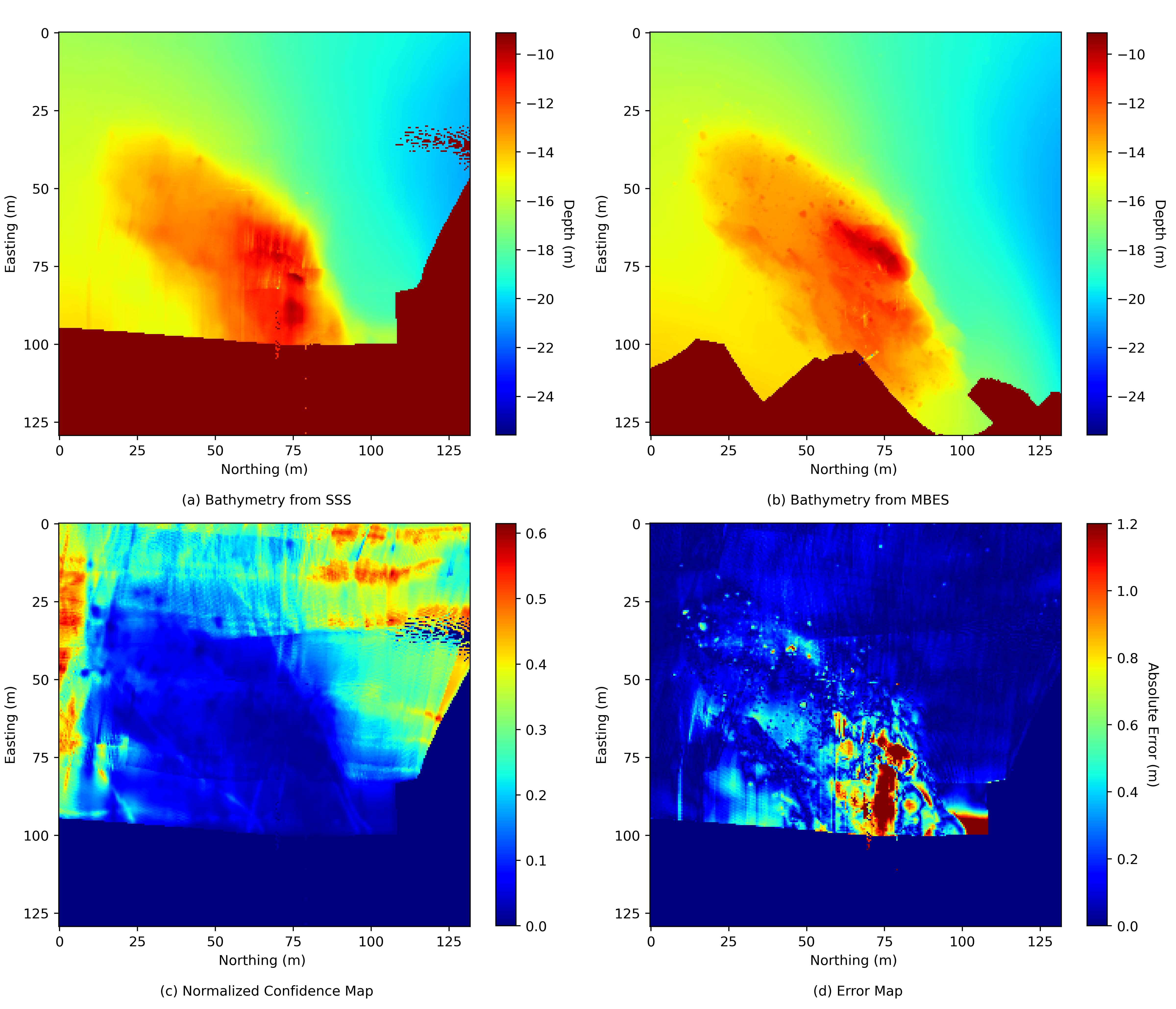}
\caption{Zoomed in section of Fig. \ref{fig:Goteborg_height_map} where the area of interest contains a hill with multiple boulders.}
\label{fig:goteborg_height_map_zoomin}
\end{figure}
Fig. \ref{fig:Goteborg_height_map} shows the reconstructed bathymetry with six sidescan survey lines on the test data from Dataset 1, with a mean absolute error 0.059 m. Looking at Fig. \ref{fig:Goteborg_height_map}(a), we observe that the reconstructed bathymetry reproduces the seafloor topography on a large scale. It also highlights one advantage of sidescan over multibeam, wider swath coverage. Only six survey lines can cover around 60\% of the surveyed area. The MBES from these six lines would only cover about 35\% of the area, indicating that the proposed method could in theory significantly improve the survey efficiency.
Fig. \ref{fig:Goteborg_height_map}(c) shows the corresponding confidence map where clearly the bottom right of the map has low confidence, high uncertainty. In Fig. \ref{fig:Goteborg_height_map}(d) we see that the areas with the highest errors have high predicted uncertainty.  In the zoomed in Fig.  \ref{fig:goteborg_height_map_zoomin}, we can observe that the low confidence area contains a huge hill with many boulders where the network's prediction performs worst. The network manages to recover the contour of the hill and some boulders, but the details are less accurate. One possible explanation is that this is an area with many topography variations, and a small error in depth prediction will cause a relatively large error in its position in the map, based on the trigonometry calculation described in  \eqref{eq:op_norm2}.

\begin{figure*}[t]
\centering
\includegraphics[width=7.0in]{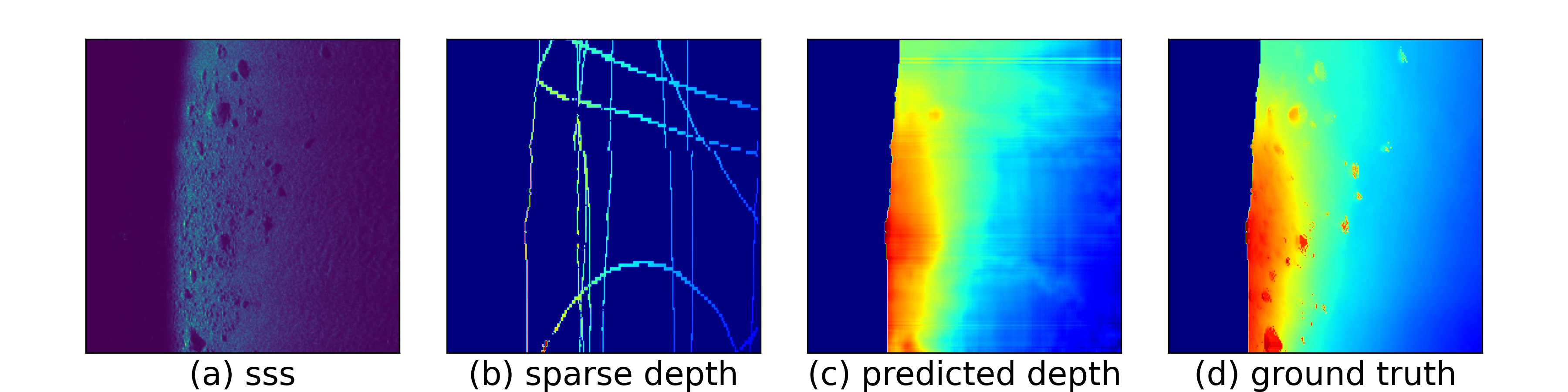}
\caption{Another example of image pairs from Dataset 1, where multiple rocks are observable from sidescan. (a) Sidescan intensities input window. (b) Sparse depth input window. (c) Predicted depth output window. (d) Ground truth depth window.}
\label{fig:windows_rocks}
\end{figure*}

One interesting finding is that, in Fig. \ref{fig:windows_rocks}, some rocks appeared in SSS are successfully recognized by the network, seen in Fig. \ref{fig:windows_rocks}(c). The network accurately infers the elevation rising in front of the shadows, which indicates the necessity and importance of sidescan data. However, not all rocks in Fig. \ref{fig:windows_rocks}(a) and (d) are shown as prominent in the prediction. This issue could possibly be addressed in the future work by adding constraints on the gradients of the sidescan intensities and increasing the sidescan's across-track resolution. 

\subsubsection{Dataset 2 and Generalization}
\begin{figure}[t]
\centering
\includegraphics[width=3.5in]{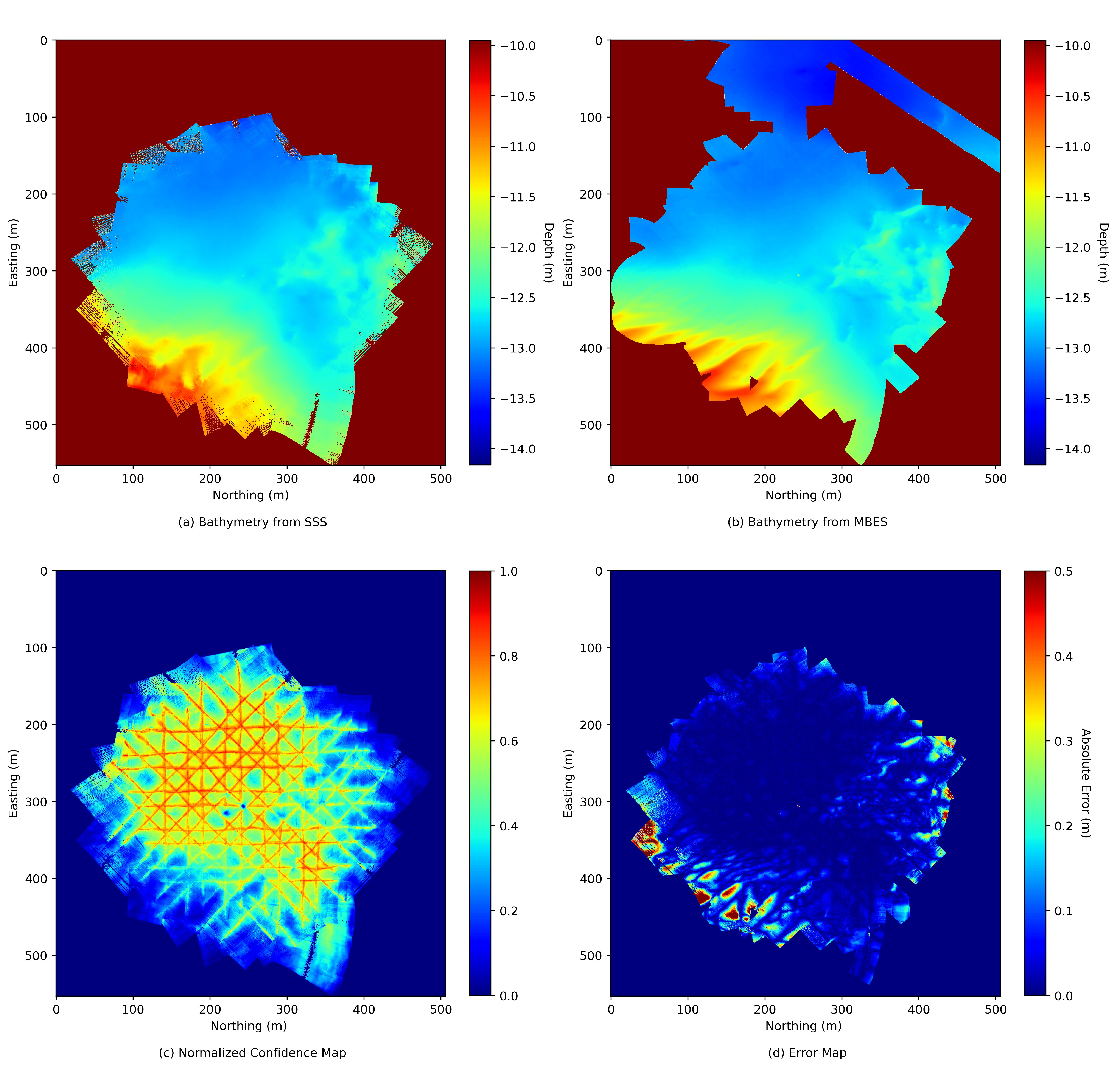}
\caption{Bathymetry on Dataset 2, produced by sidescan and multibeam. (a) Bathymetry from 36 sidescan survey lines, covering about 0.16 $km^2$ area. (b) Ground truth bathymetry produced with multibeam data. (c) Normalized Confidence map for the bathymetry produced from sidescan. (d) Absolute error map between (a) and (b).}
\label{fig:motala_height_map}
\end{figure}

\begin{figure}[t]
\centering
\includegraphics[width=3.5in]{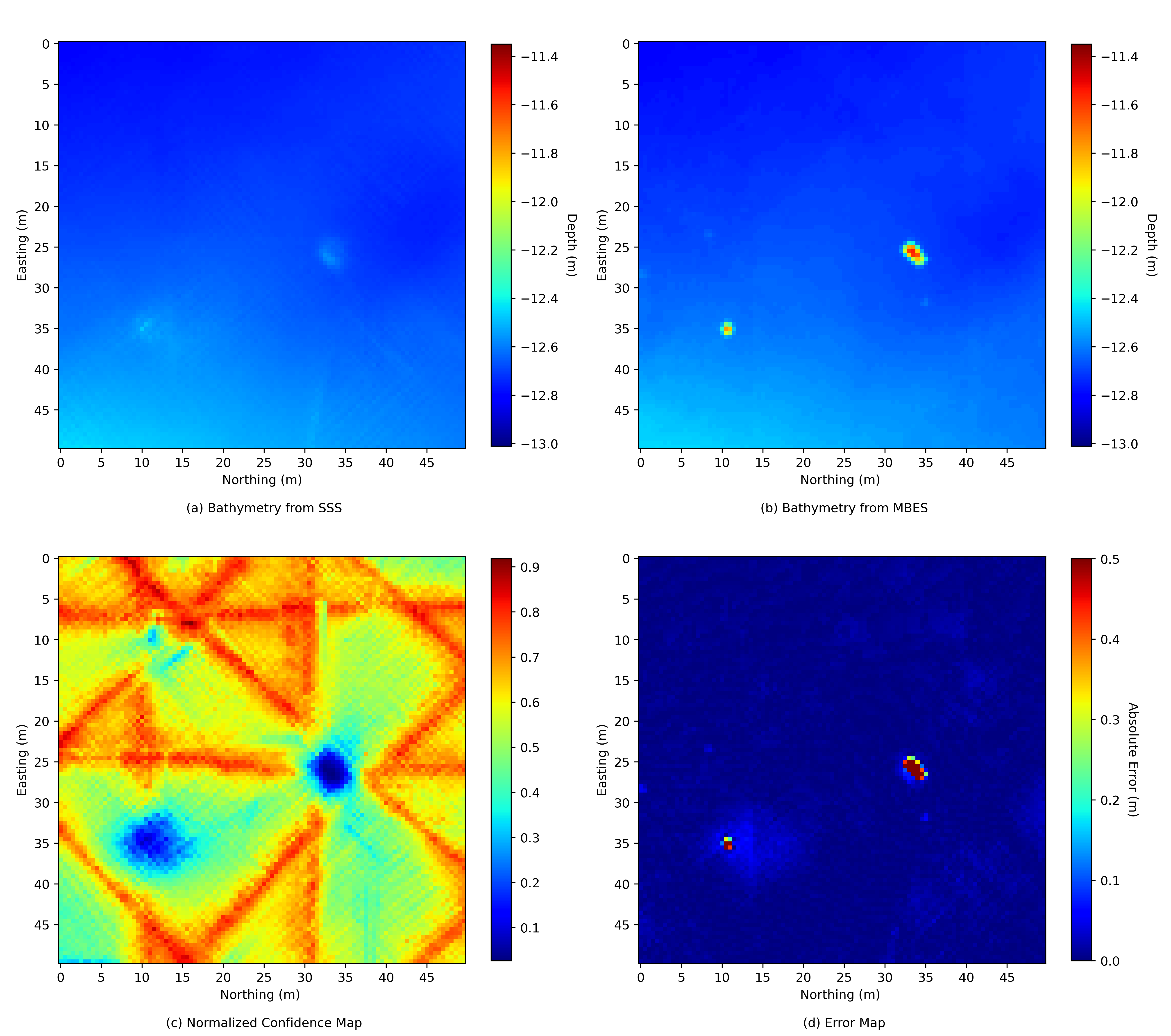}
\caption{Zoomed in section of Fig. \ref{fig:motala_height_map} where the area of interest contains two boulders.}
\label{fig:motala_height_map_zoomin}
\end{figure}

Fig. \ref{fig:motala_height_map} shows the comparison between the reconstructed bathymetry and the ground truth from Dataset 2 and the corresponding confidence map. From Fig. \ref{fig:motala_height_map}(a) and (b) we can again observe that the reconstructed bathymetry captures most of the seafloor topography. The bottom of Fig. \ref{fig:motala_height_map}(a) shows a distinguishing sidescan's characteristic that there is no measurement in the nadir area hence no estimates about the terrain. Fig. \ref{fig:motala_height_map}(c) presents the confidence map of the prediction, where we can see clearly the confidence is high along the sonar's trajectory (see Fig. \ref{fig:sparse_bathymetry}) while relatively low near to the boundary of the surveyed area. The reasons that the periphery has high uncertainty are the uncertainty of the depth estimation is naturally high as one moves away from the sonar [see Fig. \ref{fig:windows}(c)] and there is no longer sparse depth available to reduce the drifting errors. Another observation is that in the middle of the surveyed area, there are two places with sudden low confidence, where they are two boulders. If we zoom in there, as seen in From Fig. \ref{fig:motala_height_map_zoomin}(a) we can see that the two boulders are not as sharply shown as in Fig. \ref{fig:motala_height_map_zoomin}(b).

Dataset 2 covers a relatively flat area with a different depth distribution in the seafloor. The reconstructed bathymetry has a 0.043 m absolute error, indicating a good generalization ability on the unseen data of a different natural environment.   This indicates that one could train a network using SSS and MBES. Thereafter, use it on many AUVs equipped only with the same SSS. 

\subsection{Effects of Sparse Depth}
\begin{figure}[!t]
\centering
\subfloat[Corrupted Sparse Depth]{\includegraphics[width=1.75in]{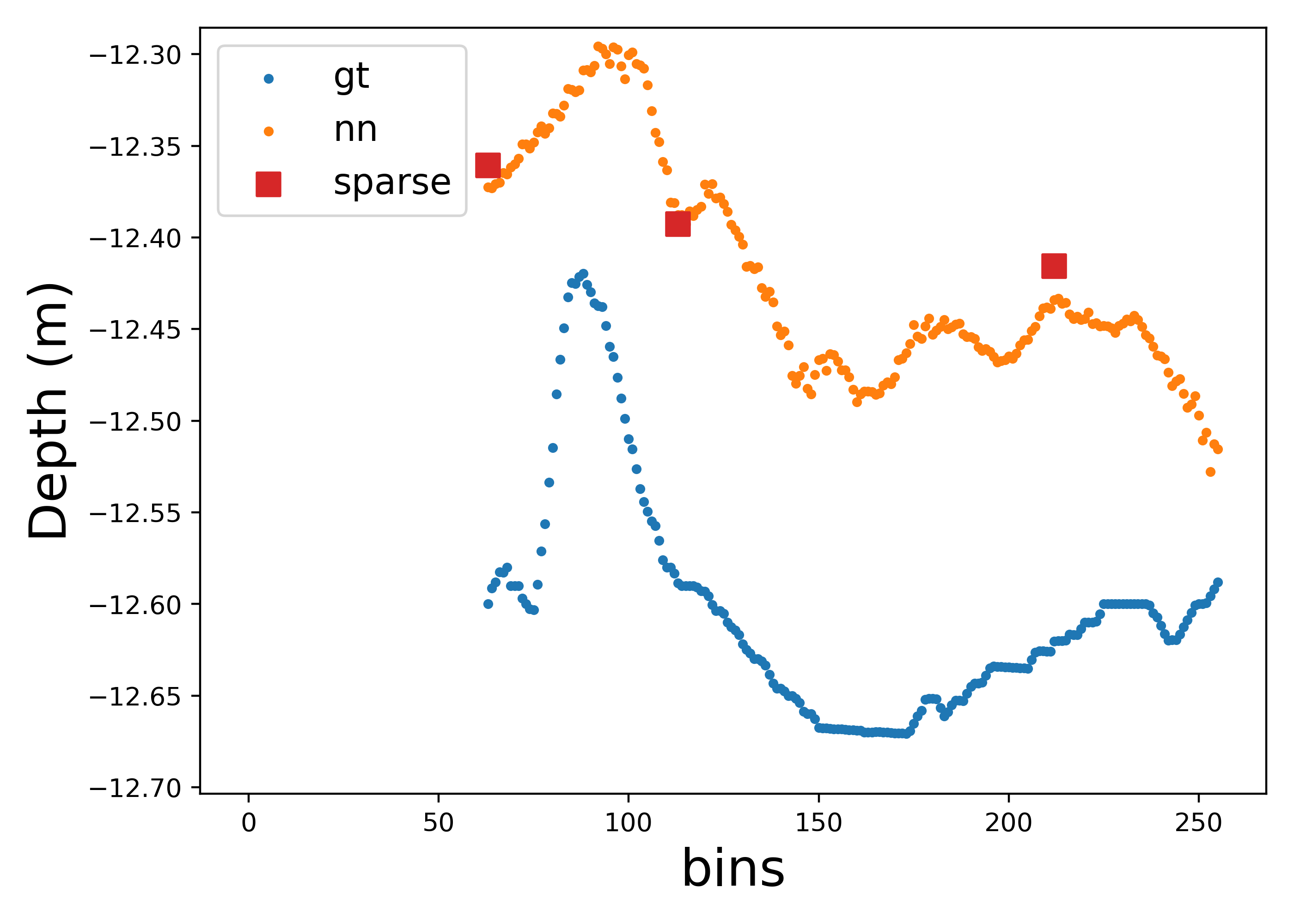}
\label{fig:sparse_depth_quality_a}}
\subfloat[Accurate Sparse Depth]{\includegraphics[width=1.75in]{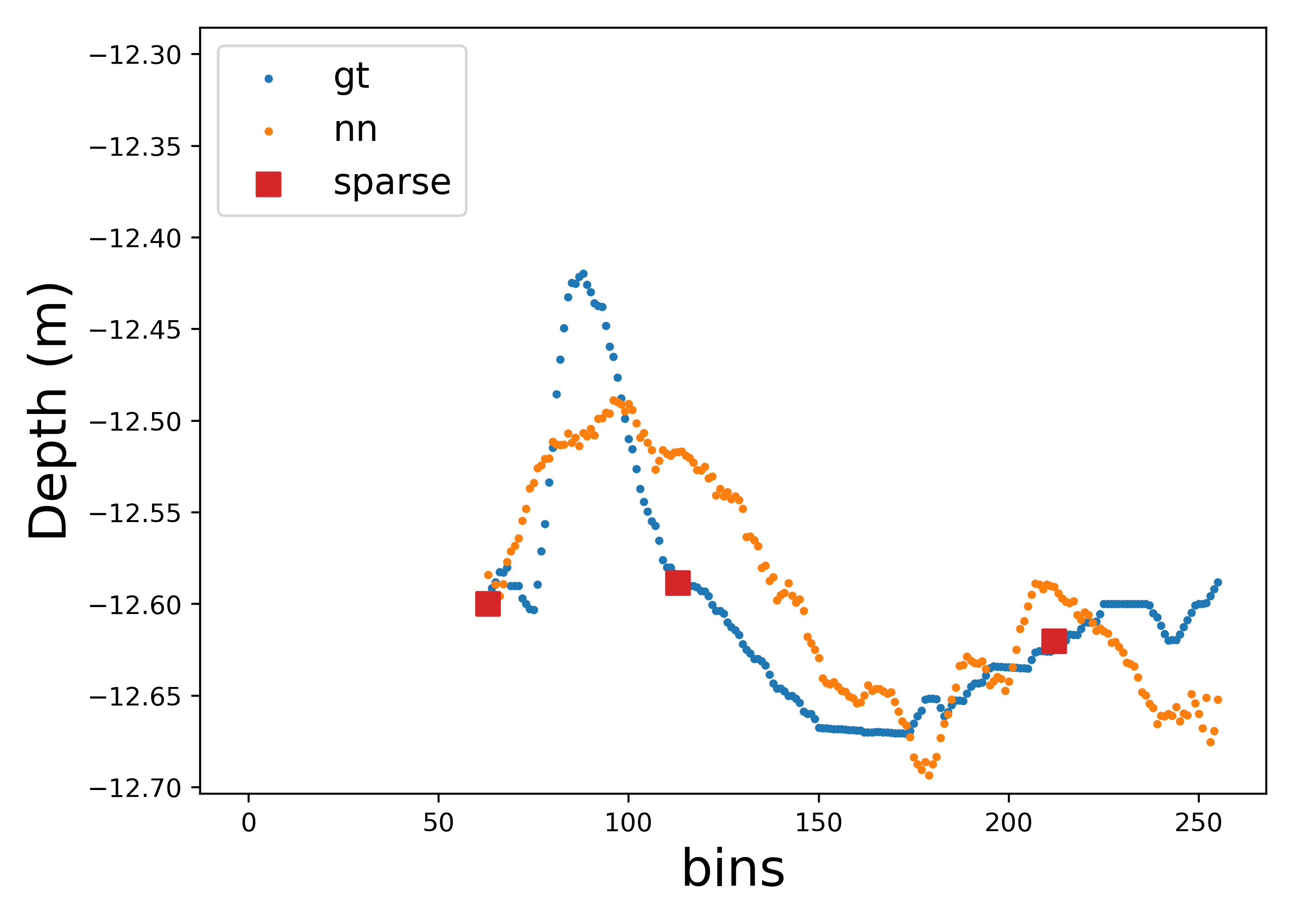}
\label{fig:sparse_depth_quality_b}}
\caption{Effects of sparse depth quality: Here we plot one row of the network's input and output; one ping of ground truth depth in blue, one ping of the depth prediction in orange, and the provided sparse depth as input in red. (a) The sparse depth provided is corrupted, thus not aligned with the ground truth depth, leading the prediction of the network off by a lot. (b) The sparse depth provided is accurate, leading the prediction aligned much better with the ground truth.}
\label{fig:sparse_depth_quality}
\end{figure}

Another interesting observation is that the quality and quantity of the sparse depth are critical for our proposed method to reconstruct a good bathymetry. The provided sparse depth acts as a boundary constraint in the optimization, so if the quality of the input sparse depth is low, i.e., the measurements being corrupted, the depth estimation will have large errors. As seen in Fig. \ref{fig:sparse_depth_quality}, when the provided sparse depth is inaccurate, the shape of the predicted depth contour is more or less right but with an offset due to the errors in sparse depth. Several reasons could cause inaccurate sparse depth measurement, e.g., errors from the altimeter sensor, affecting the quality of reconstructed bathymetry.

Not only the quality of the sparse depth but also the quantity affects the prediction accuracy. In Table \ref{tab:sparse-depth} we compare the mean absolute error on the bathymetric map generated by the neural network, with different numbers of survey lines to provide sparse depth as constraints. As we can see in the table, when all of the survey lines are utilized, the prediction accuracy is the highest, and as the quantity of provided sparse depth decreases, the prediction accuracy decreases.

In practice, for example, Dataset 2, one could certainly use less than 36 survey lines for the sidescan to cover the whole area. We showed in Table \ref{tab:sparse-depth} the reconstructed error is still relatively low even with 30\% sparse depth provided, indicating 70\% efficiency improvement. So one could carefully plan the sidescan survey to cover a much larger area within one mission to construct a high-quality bathymetric map with the proposed method.

\begin{table}[!t]
\renewcommand{\arraystretch}{1.3}
\caption{Effects of Sparse Depth quantity}
\label{tab:sparse-depth}
\centering
\begin{tabular}{c||c|c}
\hline
 & Dataset 1 & Dataset 2\\
  & Testing MAE (m) & Testing MAE (m)\\
\hline
100\% sparse depth & 0.059 & 0.043\\
\hline
50\% sparse depth & 0.074 & 0.053\\
\hline
30\% sparse depth & 0.082 & 0.060\\
\hline
no sparse depth & 0.734 & 0.200\\
\hline

\end{tabular}
\end{table}

\subsection{Effects of Uncertainty Estimation}
\begin{figure}[t]
\centering
\includegraphics[width=3.5in]{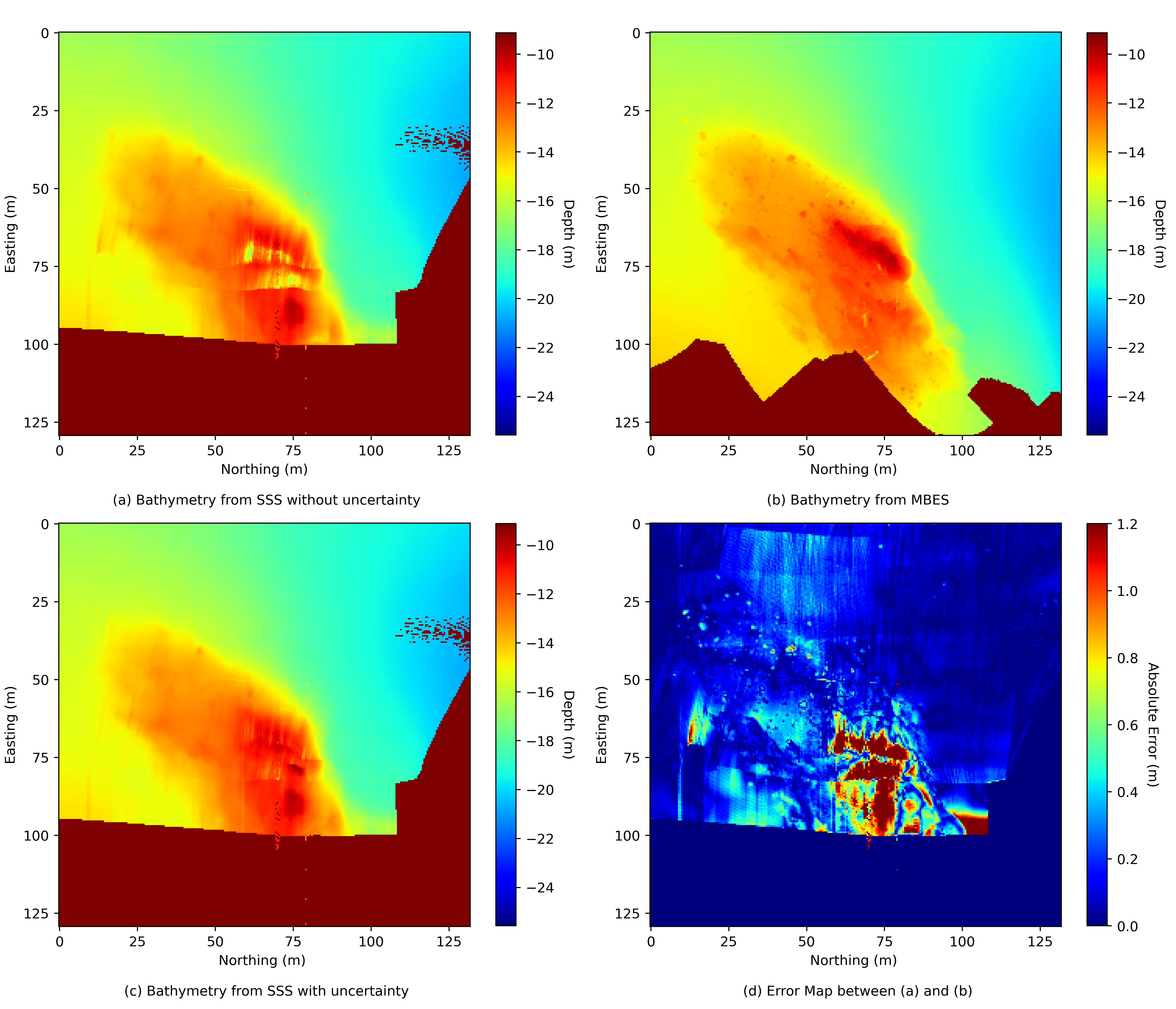}
\caption{Zoomed in section of bathymetry for Dataset 1, without using the uncertainty estimation in (a). The same Fig. \ref{fig:goteborg_height_map_zoomin}(a), bathymetry reconstructed with uncertainty in (c). }
\label{fig:goteborg_height_map_zoomin_no_uncertainty}
\end{figure}

Uncertainty estimation is useful when fusing the estimated bathymetry from each sidescan line. We use Dataset 1 to illustrate that uncertainty could improve the quality of the reconstructed bathymetry. Assuming we lack uncertainty estimation and fuse the bathymetry estimation simply averaging each estimate, we can generate a bathymetry with absolute error 0.071 m while using uncertainty as described in \eqref{equ:bathymetry_model} we can achieve 0.059 m error. Besides that, the fusion without using uncertainty performs much worse in the areas that are supposed to be highly uncertain. Fig. \ref{fig:goteborg_height_map_zoomin_no_uncertainty} shows the same place as Fig. \ref{fig:goteborg_height_map_zoomin} but without using the uncertainty estimation. We can clearly observe from Fig. \ref{fig:goteborg_height_map_zoomin_no_uncertainty}(a) that the bathymetric map is much worse when uncertainty is not used. 
\subsection{Bathymetry with Higher Resolution}
\begin{figure}[t]
\centering
\includegraphics[width=3.5in]{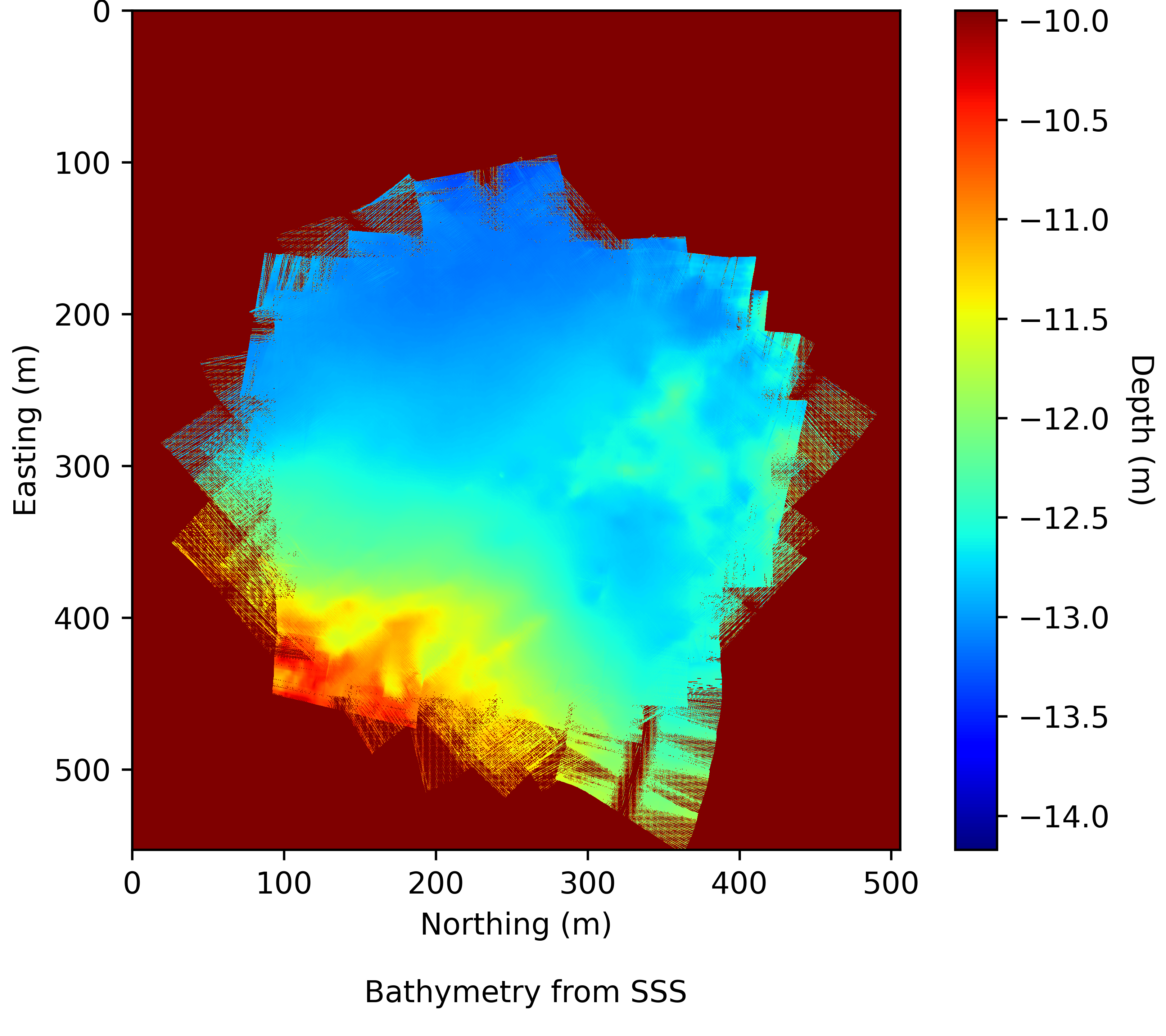}
\caption{Bathymetry reconstructed from sidescan with 0.25 m grid size for Dataset 2.}
\label{fig:motala_height_map_0.25m}
\end{figure}

When reconstructing the bathymetry from SSS we use the grid size 0.5 m because our bathymetric map from MBES (Dataset 1) has 0.5 m resolution, which is used to generate the training data. Nevertheless, for Dataset 2 we do have a bathymetric map from MBES with 0.25 m resolution, which can be used as ground truth to compare the bathymetry from sidescan with a grid size of 0.25 m, as shown in Fig. \ref{fig:motala_height_map_0.25m}. To generate such map, we use the same outputs from the neural network but only use a smaller grid size when constructing the bathymetry from predicted depth windows.
We can clearly observe the effects of ship turning on sidescan swaths from the bottom right of Fig. \ref{fig:motala_height_map_0.25m}, where the portions in the inside corners overlap while the portions in the outside corners have incomplete coverage for this fine scale grid. The absolute error compared to the ground truth is 0.042 m, very close to the error with 0.5 m grid size in Table \ref{tab:sparse-depth}. 

\section{Conclusion and future work}
We presented a novel approach to reconstruct high-resolution bathymetry from sidescan data using a neural network. The neural network is trained in an end-to-end fashion to predict the depth and uncertainty from sidescan intensities and sparse depth. The predicted depth and uncertainty, modeled as following a Laplacian distribution, are fused to construct the bathymetry. In the qualitative and quantitative analysis, we showed that the generated bathymetry has high quality and low errors below the decimeter level. We also showed the important role both the sparse depth and the confidence estimate plays on the accuracy of the fused map. 

The current network architecture generates independent depth windows from sidescan data between a fixed time period without incorporating the sequential nature of the sidescan pings. In future work, we would like to better address this by using a recurrent FCN model conditioned on the previous pings.

Another interesting direction is to use sidescan measurements with a higher across-track resolution to fully utilize the advantages of the sidescan and generate a bathymetry with a higher resolution than the one generated from the multibeam. One challenge here will be the lack of ground truth to analyze the performance quantitatively. Another challenge is that with a higher across-track resolution, the larger the width of images will be. One may use super-resolution CNN model \cite{yamanaka2017fast} to address such challenge, or one may treat the sidescan ping by ping so that higher resolution would not be too computational heavy.


%



\section*{Acknowledgment}
This work was partially supported by the Wallenberg AI, Autonomous
Systems and Software Program (WASP) and  by the Stiftelsen  för  Strategisk  Forskning (SSF)  through  the  Swedish  Maritime  Robotics  Centre  (SMaRC)(IRC15-0046). Our dataset was acquired in collaboration with MarinMätteknik (MMT) Gothenburg.

\ifCLASSOPTIONcaptionsoff
  \newpage
\fi



%
\bibliographystyle{IEEEtran}
\bibliography{IEEEabrv,ieeeref}

\begin{thebibliography}{10}
\providecommand{\url}[1]{#1}
\csname url@samestyle\endcsname
\providecommand{\newblock}{\relax}
\providecommand{\bibinfo}[2]{#2}
\providecommand{\BIBentrySTDinterwordspacing}{\spaceskip=0pt\relax}
\providecommand{\BIBentryALTinterwordstretchfactor}{4}
\providecommand{\BIBentryALTinterwordspacing}{\spaceskip=\fontdimen2\font plus
\BIBentryALTinterwordstretchfactor\fontdimen3\font minus
  \fontdimen4\font\relax}
\providecommand{\BIBforeignlanguage}[2]{{%
\expandafter\ifx\csname l@#1\endcsname\relax
\typeout{** WARNING: IEEEtran.bst: No hyphenation pattern has been}%
\typeout{** loaded for the language `#1'. Using the pattern for}%
\typeout{** the default language instead.}%
\else
\language=\csname l@#1\endcsname
\fi
#2}}
\providecommand{\BIBdecl}{\relax}
\BIBdecl

\bibitem{folkesson20}
J.~{Folkesson}, H.~{Chang}, and N.~{Bore}, ``Lambert’s cosine law and
  sidescan sonar modeling,'' in \emph{IEEE/OES Auton. Underwater Veh. Symp.},
  2020, pp. 1--6.

\bibitem{xie2019inferring}
Y.~Xie, N.~Bore, and J.~Folkesson, ``Inferring depth contours from sidescan
  sonar using convolutional neural nets,'' \emph{IET Radar, Sonar \&
  Navigation}, vol.~14, no.~2, pp. 328--334, 2019.

\bibitem{laina2016deeper}
I.~{Laina}, C.~{Rupprecht}, V.~{Belagiannis}, F.~{Tombari}, and N.~{Navab},
  ``Deeper depth prediction with fully convolutional residual networks,'' in
  \emph{Proc. Int. Conf. 3D Vision}, 2016, pp. 239--248.

\bibitem{ma2018sparse}
\BIBentryALTinterwordspacing
F.~{Ma} and S.~{Karaman}, ``Sparse-to-dense: Depth prediction from sparse depth
  samples and a single image,'' in \emph{Proc. IEEE Int. Conf. Robot. Autom.},
  2018, pp. 4796--4803. [Online]. Available:
  \url{https://doi.org/10.1109/ICRA.2018.8460184}
\BIBentrySTDinterwordspacing

\bibitem{woock2010deep}
\BIBentryALTinterwordspacing
P.~{Woock} and C.~{Frey}, ``Deep-sea auv navigation using side-scan sonar
  images and slam,'' in \emph{Proc. IEEE OCEANS Conf. Eur.}, 2010, pp. 1--8.
  [Online]. Available: \url{https://doi.org/10.1109/OCEANSSYD.2010.5603528}
\BIBentrySTDinterwordspacing

\bibitem{li1991improvement}
\BIBentryALTinterwordspacing
R.~Li and S.~Pai, ``Improvement of bathymetric data bases by shape from shading
  technique using side-scan sonar images,'' in \emph{Proc. IEEE OCEANS Conf.},
  vol.~1, Honololu, HI, USA, 1991, pp. 320--324. [Online]. Available:
  \url{https://doi.org/10.1109/OCEANS.1991.613950}
\BIBentrySTDinterwordspacing

\bibitem{zhang1999shape}
\BIBentryALTinterwordspacing
R.~Zhang, P.-S. Tsai, J.~E. {Cryer}, and M.~{Shah}, ``Shape-from-shading: a
  survey,'' \emph{{IEEE} Trans. Pattern Anal. Mach. Intell.}, vol.~21, no.~8,
  pp. 690--706, 1999. [Online]. Available:
  \url{https://doi.org/10.1109/34.784284}
\BIBentrySTDinterwordspacing

\bibitem{coiras2007multiresolution}
\BIBentryALTinterwordspacing
E.~{Coiras}, Y.~{Petillot}, and D.~M. {Lane}, ``Multiresolution 3-d
  reconstruction from side-scan sonar images,'' \emph{{IEEE} Trans. Image
  Process.}, vol.~16, no.~2, pp. 382--390, 2007. [Online]. Available:
  \url{https://doi.org/10.1109/TIP.2006.888337}
\BIBentrySTDinterwordspacing

\bibitem{jones2018method}
K.~R. Jones and P.~Traykovski, ``A method to quantify bedform height and
  asymmetry from a low-mounted sidescan sonar,'' \emph{J. Atmos. Ocean.
  Technol.}, vol.~35, no.~4, pp. 893--910, 2018.

\bibitem{burguera2016high}
\BIBentryALTinterwordspacing
A.~Burguera and G.~Oliver, ``High-resolution underwater mapping using side-scan
  sonar,'' \emph{PLOS ONE}, vol.~11, no.~1, pp. 1--41, 2016. [Online].
  Available: \url{https://doi.org/10.1371/journal.pone.0146396}
\BIBentrySTDinterwordspacing

\bibitem{zhao2018reconstructing}
\BIBentryALTinterwordspacing
J.~Zhao, X.~Shang, and H.~Zhang, ``Reconstructing seabed topography from
  side-scan sonar images with self-constraint,'' \emph{Remote Sens.}, vol.~10,
  no.~2, p. 201, 2018. [Online]. Available:
  \url{http://dx.doi.org/10.3390/rs10020201}
\BIBentrySTDinterwordspacing

\bibitem{dzieciuch2016non}
I.~Dzieciuch, D.~Gebhardt, C.~Barngrover, and K.~Parikh, ``Non-linear
  convolutional neural network for automatic detection of mine-like objects in
  sonar imagery,'' in \emph{Proc. Int. Conf. Appl. Nonlinear Dyn.}\hskip 1em
  plus 0.5em minus 0.4em\relax Springer, 2017, pp. 309--314.

\bibitem{huo2020underwater}
\BIBentryALTinterwordspacing
G.~{Huo}, Z.~{Wu}, and J.~{Li}, ``Underwater object classification in sidescan
  sonar images using deep transfer learning and semisynthetic training data,''
  \emph{IEEE Access}, vol.~8, pp. 47\,407--47\,418, 2020. [Online]. Available:
  \url{https://doi.org/10.1109/ACCESS.2020.2978880}
\BIBentrySTDinterwordspacing

\bibitem{einsidler2018deep}
\BIBentryALTinterwordspacing
D.~{Einsidler}, M.~{Dhanak}, and P.~{Beaujean}, ``A deep learning approach to
  target recognition in side-scan sonar imagery,'' in \emph{Proc. IEEE/MTS
  OCEANS Conf.}, Charleston, SC, USA, 2018, pp. 1--4. [Online]. Available:
  \url{https://doi.org/10.1109/OCEANS.2018.8604879}
\BIBentrySTDinterwordspacing

\bibitem{rahnemoonfar2019semantic}
\BIBentryALTinterwordspacing
M.~{Rahnemoonfar} and D.~{Dobbs}, ``Semantic segmentation of underwater sonar
  imagery with deep learning,'' in \emph{IEEE Int. Geosci. Remote Sens. Symp.},
  2019, pp. 9455--9458. [Online]. Available:
  \url{https://doi.org/10.1109/IGARSS.2019.8898742}
\BIBentrySTDinterwordspacing

\bibitem{wu2019ecnet}
\BIBentryALTinterwordspacing
M.~Wu \emph{et~al.}, ``Ecnet: Efficient convolutional networks for side scan
  sonar image segmentation,'' \emph{Sensors}, vol.~19, no.~9, p. 2009, 2019.
  [Online]. Available: \url{http://dx.doi.org/10.3390/s19092009}
\BIBentrySTDinterwordspacing

\bibitem{redmon2016you}
\BIBentryALTinterwordspacing
J.~{Redmon}, S.~{Divvala}, R.~{Girshick}, and A.~{Farhadi}, ``You only look
  once: Unified, real-time object detection,'' in \emph{Proc. IEEE Conf.
  Comput. Vision Pattern Recognit.}, 2016, pp. 779--788. [Online]. Available:
  \url{https://doi.org/10.1109/CVPR.2016.91}
\BIBentrySTDinterwordspacing

\bibitem{liu2015deep}
F.~Liu, C.~Shen, and G.~Lin, ``Deep convolutional neural fields for depth
  estimation from a single image,'' in \emph{Proc. IEEE Conf. Comput. Vision
  Pattern Recognit.}, 2015, pp. 5162--5170.

\bibitem{lakshminarayanan2016simple}
B.~Lakshminarayanan, A.~Pritzel, and C.~Blundell, ``Simple and scalable
  predictive uncertainty estimation using deep ensembles,'' in \emph{Proc. Int.
  Conf. Neural Inf. Process. Syst.}\hskip 1em plus 0.5em minus 0.4em\relax
  Curran Associates Inc., 2017, p. 6405–6416.

\bibitem{walz2020uncertainty}
\BIBentryALTinterwordspacing
S.~{Walz}, T.~{Gruber}, W.~{Ritter}, and K.~{Dietmayer}, ``Uncertainty depth
  estimation with gated images for 3d reconstruction,'' in \emph{Proc. IEEE
  Int. Conf. Intell. Transp. Syst.}, 2020, pp. 1--8. [Online]. Available:
  \url{https://doi.org/10.1109/ITSC45102.2020.9294571}
\BIBentrySTDinterwordspacing

\bibitem{gneiting2007strictly}
\BIBentryALTinterwordspacing
T.~Gneiting and A.~E. Raftery, ``Strictly proper scoring rules, prediction, and
  estimation,'' \emph{J. Amer. Stat. Assoc.}, vol. 102, no. 477, pp. 359--378,
  2007. [Online]. Available: \url{https://doi.org/10.1198/016214506000001437}
\BIBentrySTDinterwordspacing

\bibitem{blondel2010handbook}
\BIBentryALTinterwordspacing
P.~Blondel, \emph{The handbook of sidescan sonar}, 1st~ed.\hskip 1em plus 0.5em
  minus 0.4em\relax Springer Science \& Business Media, 2010. [Online].
  Available: \url{https://doi.org/10.1007/978-3-540-49886-5}
\BIBentrySTDinterwordspacing

\bibitem{Bore2020}
\BIBentryALTinterwordspacing
N.~{Bore} and J.~{Folkesson}, ``Modeling and simulation of sidescan using
  conditional generative adversarial network,'' \emph{{IEEE} J. Ocean. Eng.},
  vol.~46, no.~1, pp. 195--205, 2021. [Online]. Available:
  \url{https://doi.org/10.1109/JOE.2020.2980456}
\BIBentrySTDinterwordspacing

\bibitem{yamanaka2017fast}
J.~Yamanaka, S.~Kuwashima, and T.~Kurita, ``Fast and accurate image super
  resolution by deep cnn with skip connection and network in network,'' in
  \emph{Proc. Int. Conf. Neural Inf. Process.}\hskip 1em plus 0.5em minus
  0.4em\relax Springer, 2017, pp. 217--225.

\end{thebibliography}




%

\begin{IEEEbiography}[{\includegraphics[width=1in,height=1.25in,clip,keepaspectratio]{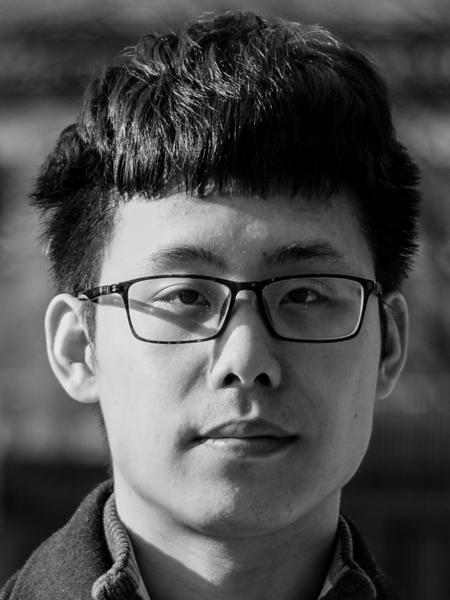}}]{Yiping Xie}
received the B.S. degree in electrical engineering from Beihang University, Beijing, China, in 2017, and the M.Sc. degree in computer science from Royal Institute of Technology (KTH), Stockholm, Sweden, in 2019. 

He is currently a Ph.D. student with the Wallenberg AI, Autonomous Systems and Software Program (WASP) from the Robotics Perception and Learning Lab at KTH. His research interests include perception for underwater robots, bathymetric mapping and localization with sidescan sonar.
\end{IEEEbiography}

\begin{IEEEbiography}[{\includegraphics[width=1in,height=1.25in,clip,keepaspectratio]{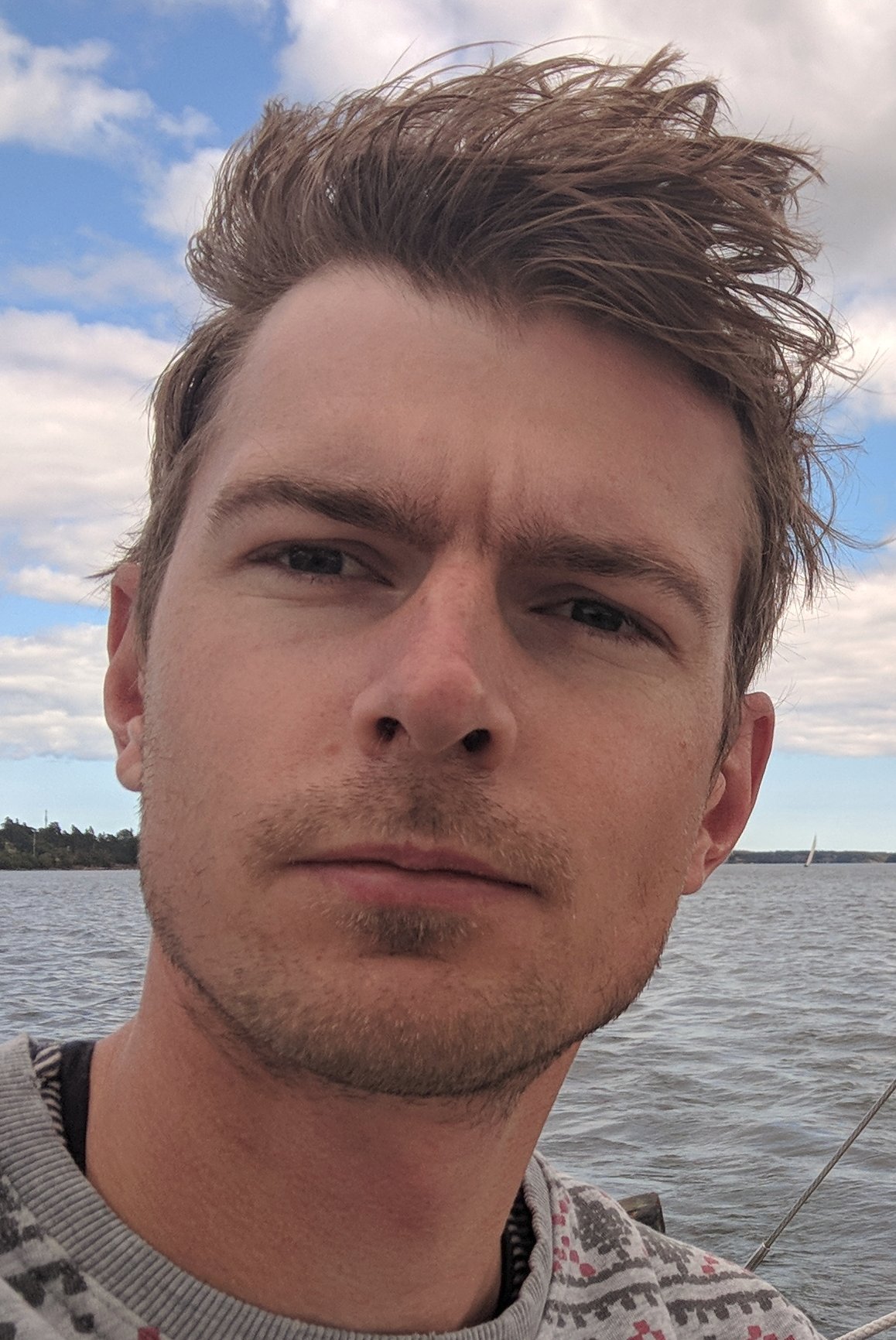}}]{Nils Bore} received the M.Sc. degree in mathematical engineering from the Faculty of Engineering, Lund University, Lund, Sweden, in 2012, and the Ph.D. degree in computer vision and robotics from the Robotics Perception and Learning Lab, Royal Institute of Technology (KTH), Stockholm, Sweden, in 2018.
He is currently a researcher with the Swedish Maritime Robotics (SMaRC) project at KTH. His research interests include robotic sensing and mapping, with a focus on probabilistic reasoning and inference. Most of his recent work has been on applications of specialized neural networks to underwater sonar data. In addition, he is interested in system integration for robust and long-term robotic deployments.
\end{IEEEbiography}


\begin{IEEEbiography}[{\includegraphics[width=1in,height=1.25in,clip,keepaspectratio]{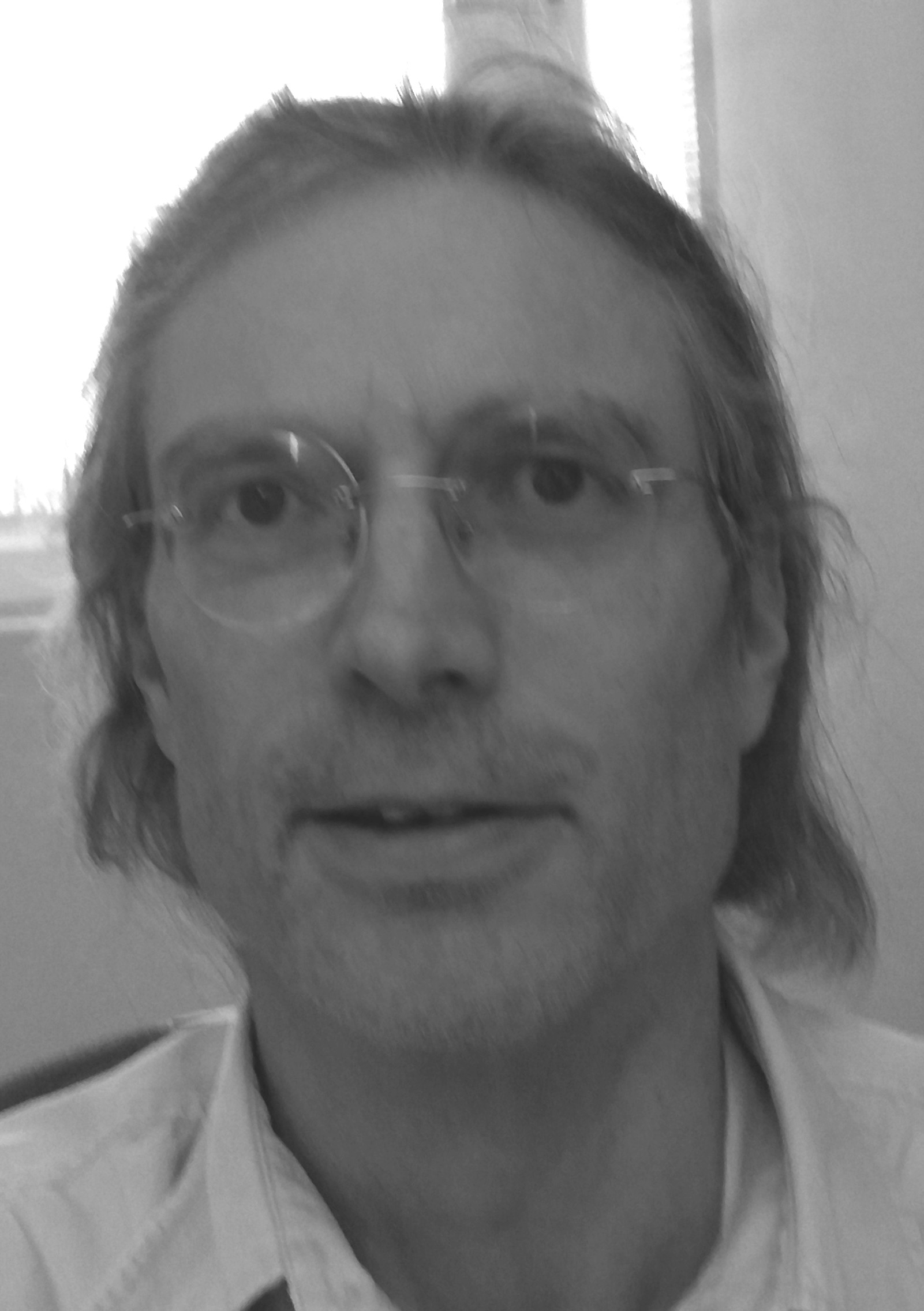}}]{John Folkesson} received the B.A. degree in physics from Queens College, City University of New York,
New York, NY, USA, in 1983, and the M.Sc. degree in computer science, and the Ph.D. degree in robotics
from Royal Institute of Technology (KTH), Stockholm, Sweden, in 2001 and 2006, respectively.
He is currently an Associate Professor of robotics with the Robotics, Perception and Learning Lab, Center for Autonomous Systems, KTH. His research interests include navigation, mapping, perception, and
situation awareness for autonomous  robots.
\end{IEEEbiography}




\end{document}